%% file: main.tex
\documentclass[letterpaper, 10 pt, conference]{ieeeconf}  

\IEEEoverridecommandlockouts                              

\usepackage{graphicx}
\usepackage{amsmath}
\usepackage{amsfonts}
\usepackage{comment}
\usepackage{xcolor}

\usepackage{url}
\usepackage{caption}
\usepackage{subcaption}

\usepackage{etoolbox}

\DeclareMathOperator*{\argmax}{argmax} 

\usepackage[ruled,linesnumbered,vlined]{algorithm2e}

\title{\LARGE \bf 
Learning Sensorimotor Primitives of Sequential Manipulation Tasks
from Visual Demonstrations}

\author{Junchi Liang, Bowen Wen, Kostas Bekris and Abdeslam Boularias
\thanks{This work was supported by NSF awards IIS 1734492 and IIS 1846043. The authors are with the Department of 
Computer Science, Rutgers University, NJ, USA. 
Emails: {\small \{junchi.liang, bowen.wen, kostas.bekris, abdeslam.boularias\}@rutgers.edu}.}%
}

\begin{document}
\maketitle


\begin{abstract}
This work aims to learn how to perform complex robot manipulation tasks that are composed of several, consecutively executed low-level sub-tasks, given as input a few visual demonstrations of the tasks performed by a person. The sub-tasks consist of moving the robot's end-effector until it reaches a sub-goal region in the task space, performing an action, and triggering the next sub-task when a pre-condition is met. Most prior work in this domain has been concerned with learning only low-level tasks, such as hitting a ball or reaching an object and grasping it. This paper describes a new neural network-based framework for learning simultaneously low-level policies as well as high-level policies, such as deciding which object to pick next or where to place it relative to other objects in the scene. A key feature of the proposed approach is that the policies are learned directly from raw videos of task demonstrations, without any manual annotation or post-processing of the data. Empirical results on object manipulation tasks with a robotic arm show that the proposed network can efficiently learn from real visual demonstrations to perform the tasks, and outperforms popular imitation learning algorithms.  
\end{abstract}



\input{introduction}

\input{related}

\input{proposed}

\input{experiment}

\section{Conclusion}
\label{sec:conclusion}
We presented a unified neural-network framework for training robots to perform complex manipulation tasks that are composed of several sub-tasks. The proposed framework employs the principal of attention by training a high-level policy network to select a pair of tool and target objects dynamically, depending on the context. 
The proposed method outperformed alternative techniques for imitation learning, without requiring any supervision beyond recorded demonstration videos.
While the current video parsing module requires the objects' CAD models beforehand, it is possible in future work to leverage model-free 6D pose trackers \cite{wen2021bundletrack} for learning from demonstration involving novel unknown objects. We will also explore other applications of the proposed framework, such as real-world assembly tasks. 

\newpage
\bibliographystyle{IEEEtran}
\bibliography{references}  

\end{document}

%% file: introduction.tex
\section{Introduction}
\label{sec:introduction}
Complex manipulation tasks are performed by combining low-level sensorimotor primitives, such as grasping, pushing and simple arm movements, with high-level reasoning skills, such as deciding which object to grasp next and where to place it. While low-level sensorimotor primitives have been extensively studied in robotics, learning how to perform high-level task planning is relatively less explored. 

\begin{figure}[t]
     \includegraphics[width=0.49\textwidth]{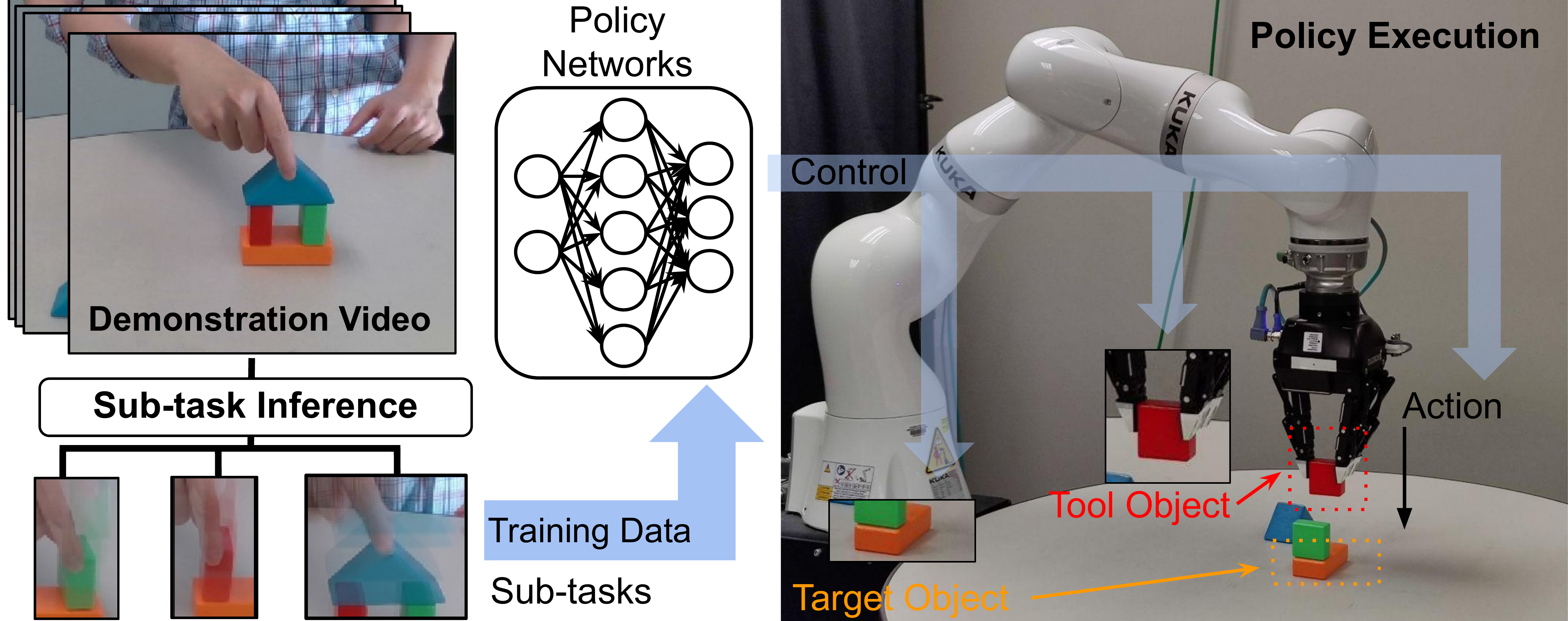}
     \vspace{-.15in}
     \caption{\small System overview: (top left) Video demonstrations of sequential manipulation tasks performed by a person. (bottom left) The manipulated objects are tracked in the input video to automatically identify sub-tasks. (center) Given the tracking, policy networks are trained to perform high-level reasoning and compute low-level controls. (right) The output of the policy networks is forwarded to a robot, which manipulates the same objects as in the demonstrations without other manual annotation. It is important to note that the objects are always {\bf randomly} placed on the table at the beginning of each demonstration and each test scenario.}
     \vspace{-.25in}
\label{fig:sys_overview}
\end{figure}

High-level reasoning consists of appropriately chaining low-level skills, such as picking and placing. It determines when the goal of a low-level skill has been reached, and the pre-conditions for switching to the next skill are satisfied. This work proposes a unified framework for learning both low and high level skills in an end-to-end manner from visual demonstrations of tasks performed by people. The focus is on tasks that require manipulating several objects in a sequence. Examples include stacking objects to form a structure, as in Fig. \ref{fig:sys_overview}, removing lug nuts from a tire to replace it, and dipping a brush into a bucket before pressing it on a surface for painting. These tasks are considered in the experimental section of this work. For all of these tasks, the pre-conditions of low-level skills depend on the types of objects as well as their spatial poses relative to each other, in addition to the history of executed actions. To support the networks responsible for the control policies, this work uses a separate vision neural network to recognize the objects and to track their 6D poses both over the demonstration videos as well as during execution. The output of the vision network is the semantic category of each object and its 6D pose relative to other objects. This output along with the history of executed actions is passed to a high-level reasoning neural network, which selects a pair of two objects that an intermediate level policy needs to focus its {\it attention} on.

The first object is referred to as {\it the tool}, and to the second one as {\it the target}. In a stacking task, the tool is the object grasped by the robot and the target is a table or any other object on top of which the tool will be placed. In the painting task, the tool is the brush and the target is the paint bucket or the canvas. If no object is grasped, then the tool is the robot's end-effector and the target is the next object that needs to be grasped or manipulated. An intermediate-level network receives the pair of objects indicated by the high-level reasoning network, their 6D poses relative to each other, and a history of executed actions. The intermediate-level network returns a sub-goal state, defined as a {\it way-point} in $SE(3)$. Finally, a low-level neural network generates the end-effector's motion to reach the way-point. The policy neural networks are summarized in Fig. \ref{fig:model}.

While the proposed formulation is not exhaustive, it allows to cast a large range of manipulation tasks, and to use the same network to learn them. The proposed architecture requires only raw RGB-D videos, without the need to segment them into sub-tasks, or even to indicate the number of sub-tasks. The efficacy of the method is demonstrated in extensive experiments using real objects in visual demonstrations, as well as both simulation and a real robot for execution.

%% file: related.tex
\section{Related Work}
\label{sec:related}	
Most of the existing techniques in imitation learning in robotics are related to learning basic low-level sensorimotor primitives, such as grasping, pushing and simple arm movements~\cite{10.5555/1795482,10.5555/3235188}.
The problem of learning spatial preconditions of manipulation skills has been addressed in some prior works~\cite{Kroemer-2016-954,Wang-2019-112298}. Random forests were used~\cite{Kroemer-2016-954} to classify configurations of pairs of objects, obtained from real images, into a family of low level skills. However, the method presented in~\cite{Kroemer-2016-954} considers only static images where the objects are in the vicinity of each other~\cite{Kroemer-2016-954}, in contrast to the proposed model, which continuously predicts low-level skills while the objects are being manipulated and moved by the robot. Moreover, it does not consider complex tasks that are composed of several low-level motor primitives ~\cite{Kroemer-2016-954}. 

A closely related line of work models each sub-task as a funnel in the state space, wherein the input and output regions of each sub-task are modeled as a multi-modal Gaussian mixture~\cite{Wang-2019-112298,DBLP:journals/ijrr/WangGKL21}, and learned from explanatory data through an elaborate clustering process.
Explicit segmentation and clustering have also been used~\cite{Su-2016-112221}.
Compared to these methods, the proposed approach is simple to reproduce and uses significantly less hyper-parameters since it does not involve any clustering process. Our approach trains an LSTM to select and remember pertinent past actions. The proposed approach also aims for data-efficiency through an  attention mechanism provided by the high-level network. Hierarchical imitation learning with high and low level policies is investigated in recent work~\cite{le2018hierarchical,shiarlis2018taco}. These methods require ground-truth labeling of each sub-task to train the high-level policy, while the proposed method is unsupervised.

Skill chaining was considered in other domains, such as 2D robot navigation~\cite{NIPS2009_e0cf1f47}. Long-horizon manipulation tasks have also been solved by using symbolic representations via Task and Motion Planning (TAMP)~\cite{DBLP:conf/ijcai/ToussaintAST19,Kaelbling93learningto,10.5555/2908515.2908520,garrett2020integrated}. Nevertheless, all the variables of the reward function in these works are assumed to be known and fully observable, in contrast to the proposed approach. A finite-state machine that supports the specification of reward functions was presented and used to accelerate reinforcement learning of structured policies~\cite{pmlr-v80-icarte18a}. In contrast to the proposed method, the structure of the reward machine was assumed to be known. A similar idea has also been adopted in other efforts~\cite{toro2019learning, camacho2019ltl}.

While 6D poses and labels of objects are provided from a vision module~\cite{wen2020se} in the proposed approach, other recent works have shown that complex tasks can be completed by learning directly from pixels~\cite{pmlr-v87-kalashnikov18a,fox2018parametrized,journals/corr/abs-1710-01813,DBLP:journals/corr/abs-1807-03480,DBLP:conf/icra/NairBFLK20,10.5555/3295222.3295258,DBLP:journals/corr/abs-1909-05829}. This objective is typically accomplished by using compositional policy structures that are learned by imitation~\cite{pmlr-v87-kalashnikov18a,fox2018parametrized}, or that are manually specified~\cite{journals/corr/abs-1710-01813,DBLP:journals/corr/abs-1807-03480}. Some of these methods have been used for simulated control tasks~\cite{10.5555/3298483.3298491,10.5555/3327144.3327250,eysenbach2018diversity}. These promising end-to-end techniques still require orders of magnitude more training trajectories compared to methods like the one proposed here, which separates the object tracking and policy learning problems.

\begin{figure*}
     \includegraphics[width=0.95\textwidth]{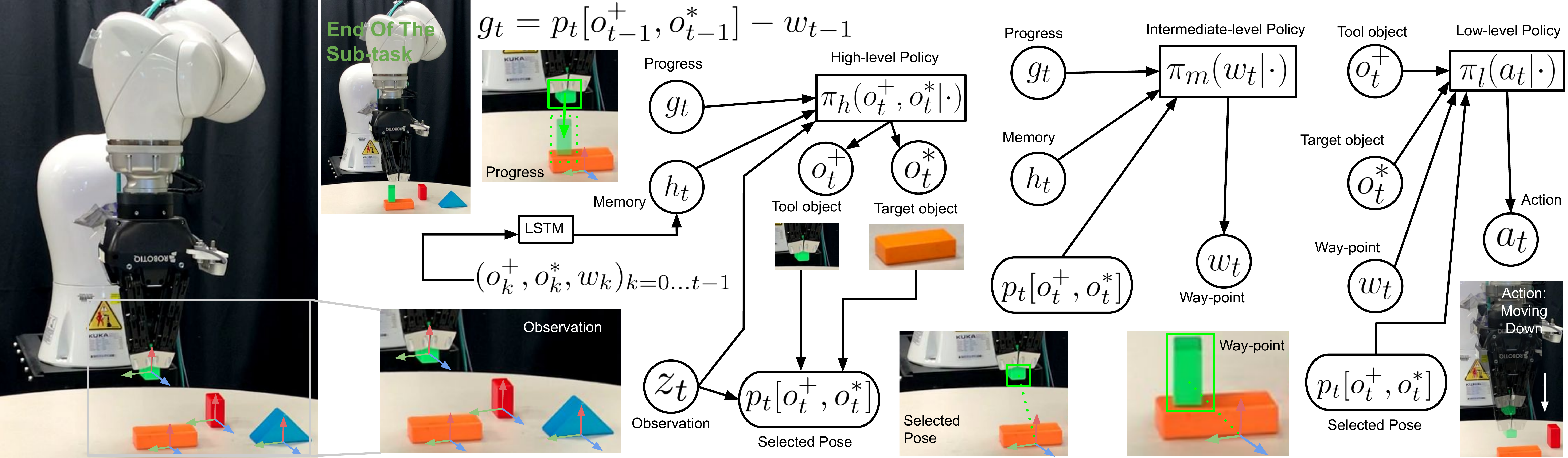}
     \caption{Left: robot performing a stacking task. Right: policy networks. A high-level policy network uses the latest observation, current progress towards the current way-point and memory to select a pair of tool/target objects and their relative poses, which are then passed along to the intermediate policy for generating a next way-point. The low-level policy is responsible of generating the actual motion of the end-effector toward the way-point.}
     \vspace{-.3in}
\label{fig:model}
\end{figure*}

%% file: proposed.tex

\section{Problem Formulation and Architecture}
\label{sec:model}
This approach employs a hierarchical neural network for learning to perform tasks that require consecutive manipulation of multiple objects. The assumption is that each scene contains at most $n$ objects from a predefined set $\mathcal O =\{o^1,o^2, \dots, o^n\}$. The robot's end-effector is included as a special object in $\mathcal O$. The robot receives as inputs at each time-step $t$ sensory data as an observation $z_t = (e_t, \langle l_t^1, \dots, l_t^m \rangle, p_t)$, where $e_t\in \mathbb{R}^3\times \mathbb{SO}(3)$ is the 6D pose of the end-effector in the world frame, $m$ is the maximum number of objects present in the scene, $l_t^i$ is the semantic label of object $o^i$, and $p_t$ is a $7n\times7n$ matrix that contains the  6D poses of all objects relative to each other, i.e., $p_t[o^i,o^j]$ is a $7$-dim. vector that represents $o^i$'s orientation and translation in the frame of object $o^j$. The objects have known geometric models and have fixed frames of reference defined by their centers and 3 principal axes. The objects are detected and tracked using the technique presented in Section~\ref{sec:tracking}. The maximum number of objects $n$ is fixed a priori. 

The system returns at each time-step $t$ an action $a_t  \in \mathbb{R}^3\times \mathbb{SO}(3)$, i.e., a desired change in the pose $e_t$ of the robotic end-effector. An individual low-level sub-task is identified by a {\it tool} denoted by $o_t^+$ and a {\it target} denoted by $o_t^*$, along with a way-point $w_t$. The tool is the object being grasped by the robot at time $t$, the target is the object to manipulate using the grasped tool and the predicted way-point is the desired pose of the tool in the target's frame at the end of the sub-task. The way-point $w_t$ is a function of time as it changes based on the current pose of the tool relative to the target. Several way-points are often necessary to perform even simple tasks. For instance, in painting, a brush is the tool $o_t^+$ and a paint bucket is the target $o_t^*$. To load a brush with paint, several way-points in the bucket's frame need to be predicted. The first way-point can be when the brush touches the paint, while the second way-point is slightly above the paint. The tool $o_t^+$ and target assignment $o_t^*$ are also functions of time $t$, and change as the system switches from one sub-task to the next, based on the current observation $z_t$ and on what has been accomplished so far. For instance, after loading the brush, the robot switches to the next sub-task wherein the brush is still the tool object, but the painting canvas or surface becomes the new target object. 

In the proposed model, observations are limited to $6D$ poses of objects and their semantic labels. These observations are often insufficient by themselves for determining the current stage of the task, for deciding to terminate the current sub-task and for selecting the next sub-task. For instance, in the painting example, the vision module does not provide information regarding the current status of the brush. Therefore, the robot needs to {\it remember} whether it has already dipped the brush in the paint. Since it is not practical to keep the entire sequence of past actions in memory, the approach uses a {\it Long Short-Term Memory} (LSTM) to compress the history $h_t$ of the actions that the robot has performed so far, and use it as an input to the system along with observation $z_t$. The LSTM is trained along with the other parameters of the neural network.

The following describes the three levels of the hierarchical network architecture as depicted in Figure~\ref{fig:model}.

A {\bf high-level} policy, denoted by $\pi_{h}$, returns a probability distribution over pairs $(o^+_t,o^*_t)$ of objects, wherein $o^+_t$ is the predicted tool at time $t$, and $o^*_t$ is the predicted target at time $t$. The high-level policy takes as inputs 6D poses of objects and their semantic labels, along with the pose of the robot's end-effector. Additionally, the high-level policy $\pi_{h}$ receives as inputs a history $h_t = (o^+_k, o^*_k, w_k)_{k=0,\dots,t-1}$ of pairs of tools and targets $(o^+_k, o^*_k)$ and way-points $w_k$ at different times $k$ in the past, compressed into an LSTM unit, as well as a progress vector  $g_t=p_t[o^+_{t-1}, o^*_{t-1}]-w_{t-1}$ that indicates how far is the tool $o^+_{t-1}$ from the previous desired way-point $w_{t-1}$ with respect to the target $o^*_{t-1}$.

An {\bf intermediate-level} policy, denoted by $\pi_{m}$, receives as inputs the current tool $o^+_{t}$ and target $o^*_{t}$,
the pose $p_t[o^+_{t}, o^*_{t}]$ of $o^+_{t}$ relative to $o^*_{t}$, in addition to history $h_t$ and progress vector $g_t$. Both tool $o^+_{t}$ and target $o^*_{t}$ are predicted by the high-level policy $\pi_{h}$, as explained above. The intermediate-level policy returns a way-point $w_t\in \mathbb{R}^3\times \mathbb{SO}(3)$, expressed in the coordinates system of the target object $o^*_{t}$.

A {\bf low-level} policy, denoted by $\pi_{l}$, receives as inputs the current pose $p_t[o^+_{t}, o^*_{t}]$ of the current tool $o^+_{t}$ relative to the current target $o^*_{t}$, in addition to the way-point $w_t$ predicted by the intermediate policy, and returns a Gaussian distribution on action $a_t  \in \mathbb{R}^3\times \mathbb{SO}(3)$ that corresponds to a desired change in the pose $e_t$ of the robotic end-effector.

\section{Learning Approach}
\label{sec:learning}

In the proposed framework, an RGB-D camera is used to record a human performing an object manipulation task multiple times with varying initial placements of the objects. The pose estimation and tracking technique, explained in Section~\ref{sec:tracking}, is then used to extract several trajectories of the form $\tau = (z_1,z_2,\dots,z_H)$, wherein $z_t = (e_t, \langle l_t^1, \dots, l_t^n \rangle, p_t)$ is the observed $6D$ poses of all objects at time $t$, including the end-effector's pose $e_t$. The goal of the learning process is to learn parameters of the three policy neural networks $\pi_h,\pi_m$ and $\pi_l$ that maximize the likelihood of the data $\tau$ and the inferred way-points, tools and targets, so that the system can generalize to novel placements of the objects that did not occur in the demonstrations. The likelihood is given by:
\vspace{-.05in}
{\small
\begin{eqnarray*}
P\big((z_t,w_t,o^+_t,o^*_t)_{t=0:H} \big) = \Pi_{t=1}^{H-1} P_{A,t}P_{B,t} \text{, with}
\end{eqnarray*}
\vspace{-.1in}
\begin{equation*}
    \begin{cases}
      P_{A,t} &\overset{\Delta}{=}  P(w_t, o^+_t, o^*_t | z_{t})   \\ 
              &= \pi_{h}(o^+_t, o^*_t | z_{t},  h_t,g_t)\pi_m(w_t|h_t,g_t, p_t[o^+_t, o^*_t]),\\ 
      P_{B,t} &\overset{\Delta}{=}  P(z_{t+1} | z_{t}, w_t, o^+_t, o^*_t) \\
      &= \pi_l(e_{t+1} - e_{t}| w_t, o^+_t, o^*_t,p_t[o^+_t, o^*_t]),
    \end{cases}      
    \vspace{-.05in}
\end{equation*}
}
\noindent wherein $h_t = (w_t, o^+_t, o^*_t)_{i=0}^{t-1}$ is the history and $g_t=p_t[o^+_{t-1}, o^*_{t-1}]-w_{t-1}$ is the progress vector. 

The principal challenge here lies in the fact that the sequence $(w_t, o^+_t, o^*_t)_{t=0}^{H}$ of way-points, tools, and targets is unknown, since the proposed approach uses as inputs only 6D poses of objects at different time-steps and does not require any sort of manual annotation of the data. 

To address this problem, an iterative learning process performed in three steps is proposed. First, the low-level policy is initialized by training on basic {\it reaching} tasks. The intermediate and high-level policies are initialized with prior distributions that simply encourage time continuity and proximity of way-points to target objects. Then, an expectation-maximization (EM) algorithm is devised to infer the most likely sequence $(w_t, o^+_t, o^*_t)_{t=0}^{H}$ of way-points, tools and targets in the demonstration data $(z_t)_{t=0}^{H}$. Finally, the three policy networks are trained by maximizing the likelihood of the demonstration data $(z_t)_{t=0}^{H}$ and the pseudo ground-truth data $(w_t, o^+_t, o^*_t)_{t=0}^{H}$ obtained from the EM algorithm. This process is repeated until the inferred pseudo ground-truth data $(w_t, o^+_t, o^*_t)_{t=0}^{H}$ become constant across iterations.  

\subsection{Prior Initialization}
\label{sec:initialization}


This section first explains how the low-level policy $\pi_l$ is initialized. The most basic low-level skill is moving the end-effector between two points in $\mathbb{R}^3\times \mathbb{SO}(3)$ that are relatively close to each other. We therefore initialize the low-level policy by training the policy network, using gradient-ascent, to maximize the likelihood of straight-line movements between consecutive poses $e_{t+1}$ and $e_{t}$ of the end-effector while aiming at way-points $\hat{w}_t\overset{\Delta}{=}p_{t+1}[o^+_t, o^*_t]$. Therefore, the objective of the initialization process is given as $\max_{\theta_l} \sum_{t=1}^{H}\pi_l( e_{t+1} - e_t  | o^+_t, o^*_t, p_t [o^+_t, o^*_t] , \hat{w}_{t} )$, wherein each $\hat{w}_{t}$ is expressed in the frame of the target $o^*_t$, and $\theta_l$ are the parameters of the neural network $\pi_l$. Both $o^+_t$ and $o^*_t$ are also chosen randomly in this initialization phase. The goal is to learn simple reaching skills, which will be refined and adapted in the learning steps to produce more complex motions, such as rotations. 

The intermediate policy $\pi_m$ is responsible for selecting way-point $w_t$ given history $h_t$. It is initialized by constructing a discrete probability distribution over points $(\hat{w}_{t},\hat{w}_{t+1},\hat{w}_{t+2},\dots,\hat{w}_H)$, defined as $\hat{w}_t\overset{\Delta}{=}p_{t+1}[o^+_t, o^*_t]$. Poses $p_{t+1}$ used as way-points $\hat{w}_t$ are obtained directly from demonstration data $(z_t)_{t=0}^{H}$.
Specifically, we set $\pi_{m}(\hat{w}_k|h_t,o^+_t, o^*_t,p_t[o^+_t, o^*_t]) = 0$ for $k<t$, $\pi_{m}(\hat{w}_k|h_t,o^+_t, o^*_t,p_t[o^+_t, o^*_t]) \propto \exp{(- \alpha \|\hat{w}_k\|_2)}$ for $k=t$, and $\pi_{m}(\hat{w}_k|h_t,o^+_t, o^*_t,p_t[o^+_t, o^*_t]) \propto \exp{(- \alpha \|\hat{w}_k\|_2)} \frac{1-\beta}{H-t}$ for $k>t$, where $\alpha$ and $\beta$ are predefined fixed hyper-parameters, and $\hat{w}_k$ is expressed in the coordinates system of the target $o^*_t$. This distribution encourages way-points to be close to the target at time $t$. This distribution is constructed for each candidate target $o^*_t\in\mathcal{O}$ at each time-step $t$, except for the robot's end-effector, which cannot be a target.

High-level policy $\pi_h$ is responsible for selecting tools and targets $(o^+_t,o^*_t)$ as a function of context. It is initialized by setting the tool as the object with the most motion relative to others: $o^+_t=\arg\max_{o^i\in\mathcal{O} \setminus \{o^e\}} \sum_{o^j\in\mathcal {O}} \|p_{t+1}[o^i,o^j] - p_{t}[o^i,o^j]\|$, excluding the end-effector (or human hand) $o^e$. If all the objects besides the end-effector are stationary relative to each other, then no object is being used, and the end-effector is selected as the tool. Once the tool $o^+_t$ is fixed, the prior distribution on the target $o^+_t$ is set as: $\pi_{h}(o^+_t,o^*_t|h_t,p_t,g_t) = 0$ if $o^+_t = o^e$ (the end-effector cannot be a target), $\pi_{h}(o^+_t,o^*_t|h_t,p_t,g_t) = \gamma$ if $o^+_t = o^+_{t-1}$, and $\pi_{h}(o^+_t,o^*_t|h_t,p_t,g_t) = \frac{1-\gamma}{n-2}$ if $o^+_t \neq o^+_{t-1}$, where $o^+_{t-1}$ is obtained from history $h_t$, $n$ is the number of objects and $\gamma$ is a fixed hyper-parameter, set to a value close to $1$ to ensure that switching between targets does not occur frequently in a given trajectory.

\subsection{Pseudo Ground-Truth Inference}
After initializing $\pi_h, \pi_m$ and $\pi_l$ as in Section~\ref{sec:initialization}, the next step consists of inferring from the demonstrations $(z_t)_{t=1}^{H}$ a sequence $(o^+_t,o^*_t,w_t)_{t=1}^{H}$ of tools, targets and way-points that has the highest joint probability $P\big((z_t,w_t,o^+_t,o^*_t)_{t=0:H} \big)$ (Algorithm~\ref{algo:main}, lines 2-15). This problem is solved by using the {\it Viterbi} technique. In a forward pass (lines 2-12), the method computes the probability of the most likely sequence up to time $t-1$ that results in a choice $(o^+_t,o^*_t,w_t)$ at time $t$. The log of this probability, denoted by $F_t[(o^+_t,o^*_t,w_t)]$, is computed by taking the product of three probabilities: (i) $\pi_{h}$: the probability of switching from  $o^+_{t-1}$ and $o^*_{t-1}$ as tools and targets to $o^+_{t}$ and $o^*_{t}$, given the progress vector $g_t$ and the object poses relative to each other provided by the matrix $p_t$ (which is obtained from observation $z_t$); (ii) $\pi_{m}$: the probability of selecting as a way-point a future pose $p_{k}[o^+_t,o^*_t]$ (denoted as $w_k$, $k\geq t$) for the tool relative to the target in the demonstration trajectory; this probability is also conditioned on choices made at the previous time step $t-1$; (iii) the likelihood of the observed movement of the objects at time $t$ in the demonstration, given the choice $(o^+_t,o^*_t,w_k)$ and the relative poses of the objects with respect to each other (given by matrix $p_t$). For each candidate $(o^+_t,o^*_t,w_k)$ at time $t$, we keep in $R_t$ the trace of the candidate at time $t-1$ that maximizes their joint probability.  The backward pass (lines 13-15) finds the most likely sequence $(o^+_t,o^*_t,w_t)_{t=1}^{H}$ by starting from the end of the demonstration and following the trace of that sequence in $R_t$. The last step is to train $\pi_m$, $\pi_l$ and $\pi_h$ using the most likely sequence $(o^+_t,o^*_t,w_t)_{t=1}^{H}$ as a pseudo ground-truth for the tools, targets and way-points.


\makeatletter
\patchcmd{\@algocf@start}
  {-1.5em}
  {0pt}
  {}{}
\makeatother
\begin{algorithm}
\small
  \SetKwFunction{KwMain}{Main}
  \KwIn{
  A set of $n$ objects $\mathcal O =\{o^1,o^2, \dots, o^n\}$;
  one or several demonstration trajectories $\{z_t\}_{t=1}^{H}$, wherein $z_t = (e_t, \langle l_t^1, \dots, l_t^n \rangle, p_t)$, $e_t$ is the end-effector's pose at time $t$, $p_t[o^i,o^j]$ is the 6D pose of $o^i\in \mathcal{O}$ relative to $o^j\in \mathcal{O}$, $\forall (o^i,o^j)\in \mathcal{O} \times \mathcal{O}$;
  }
  \KwOut{High-level, intermediate-level, and low-level policies $\pi_{h}$, $\pi_{m}$ and $\pi_{l}$;}
  Initialize $\pi_{h}$, $\pi_{m}$ and $\pi_{l}$ (Section~\ref{sec:initialization});
   $F_0[ : ] \leftarrow -\infty$\;
  \For{$t=1; t \leq H; t \leftarrow t+1$ \label{marker}} {
    \ForEach{$(o^+_t,o^*_t,k) \in \mathcal{O}\times\mathcal{O}\times\{t,{t+1},\dots,H\}$}
        {
        $x_t \overset{\Delta}{=} (o^+_t,o^*_t,k)$; $w_k \overset{\Delta}{=}  p_{k}[o^+_t,o^*_t]$;\\
        $\Delta p_t \overset{\Delta}{=}  p_{t+1}[o^+_t, o^*_t] - p_t[o^+_t, o^*_t]$;\\
        \ForEach{$(o^+_{t-1},o^*_{t-1},k') \in \mathcal{O}\times\mathcal{O}\times\{t-1,\dots,H\}$}
            {
            $x_{t-1} \overset{\Delta}{=} (o^+_{t-1},o^*_{t-1},k')$;
            $w_{k'} \overset{\Delta}{=}  p_{k'}[o^+_{t-1},o^*_{t-1}]$;\\
            $h_t\leftarrow (o^+_{t-1},o^*_{t-1}, w_{k'})$;\\
            $g_t \leftarrow  p_{t}[o^+_t, o^*_t] - w_t$;\\
            $Q[x_{t-1},x_{t}] \leftarrow \log\big( \pi_{h}(o^+_t,o^*_t|h_t,p_t,g_t)\big)+\log\big(\pi_{m}(w_k|h_t,o^+_t, o^*_t,p_t[o^+_t, o^*_t])\big) +\log \big(\pi_l(  \Delta p_t [o^+_t, o^*_t] | o^+_t, o^*_t, p_t [o^+_t, o^*_t] , w_k )\big) + F_{t-1}[x_{t-1}] $;
            }
            $F_t[x_t] \leftarrow \max_{x_{t-1}} Q[x_{t-1},x_{t}]$;\\
            $R_t[x_t] \leftarrow \arg\max_{x_{t-1}} Q[x_{t-1},x_{t}]$;\\
        }
  }
  \tcc{Construct the most likely sequence}
  $(o^+_H,o^*_H,k)\leftarrow \argmax_{x} R_t[x]$; $w_H\leftarrow p_{k}[o^+_H,o^*_H]$;\\
  \For{$t=H-1; t>0; t \leftarrow t-1$}{
  $(o^+_{t},o^*_{t},k)\leftarrow \argmax_{x} R_{t+1}[x]$; $w_t\leftarrow p_{k}[o^+_t,o^*_t]$;\\
  }
  Train the policy networks $\pi_{h}$, $\pi_{m}$ and $\pi_{l}$ with $(o^+_t,o^*_t,w_t)_{t=1}^{H}$ and $\{z_t\}_{t=1}^{H}$;\\
  Optional:  Go to \ref{marker} and repeat with updated policies $\pi_{h}$, $\pi_{m}$, $\pi_{l}$;
\caption{\small Learning Policies  from Visual Demonstrations}
\label{algo:main}
\end{algorithm}

\subsection{Training the Policy Networks}
To train $\pi_m$, $\pi_l$ and $\pi_h$ using the pseudo ground-truth $(o^+_t,o^*_t,w_t)_{t=1}^{H}$, obtained as explained in the previous section, we apply the stochastic gradient-descent technique to simultaneously optimize the parameters of the networks by minimizing a loss function $L$ defined as follows. 
$L$ is defined as the sum of multiple terms. The first two are 
 $L_{o^+} = \sum_{t} CE(\hat{o^+}_t, o^+_t)$ and $L_{o^*}=\sum_{t} CE(\hat{o^*}_t, o^*_t)$ where $CE$ is the cross entropy, and $(\hat{o^+}_t,\hat{o^*}_t)$ is the current prediction of $\pi_h$. The third term is $L_{w}=\sum_{t} MSE(\hat{w}_t, w_t)$ where $\hat{w}_t$ is the output of $\pi_m$ and $MSE$ is the mean square error. The next term is $L_{action}=-log[\pi_l(a_t| p_t[o^+_t, o^*_t], w_t, o^+_t, o^*_t)]$, which corresponds to the log-likelihood of the low-level actions in the demonstrations.  To further facilitate the training, two auxiliary losses are introduced. The first one is to encourage consistency within each sub-task. As the role of the memory in the architecture is to indicate the sequence sub-tasks that have been already performed, it should not change before $(o^+_t,o^*_t,w_t)$ changes. If we denote the LSTM's output as $M(h_t)$ at time-step $t$, the consistency loss is defined as  $L_{mem}=\sum_{t}\| M(h_t) - M(h_{t-1}) \|_2 \times I_{o^+_{t-1}=o^+_{t-1}, o^*_{t-1}=o^*_t,w_{t-1}=w_t}$ where $I$ is the indicator function. The last loss term, $L_{ret}$,  is used to ensure that memory $M(h_t)$ retains sufficient information from previous steps. Thus, we train an additional layer $(ret_{o^*},ret_{o^+}, ret_{w})$ directly after $M(h_t)$ to retrieve at step $t$ the target, tool and way-point of step $t-1$. $L_{ret}$ is defined as $L_{ret}=CE(ret_{o^*}(M(h_t)), o^*_{t-1})+CE(ret_{o^+}(M(h_t)), o^+_{t-1})+MSE(ret_{w}(M(h_t)), w_{t-1}).$ As a result, the complete proposed architecture  is trained with the loss $L=L_{o^+} + L_{o^*} + L_{w} + L_{action} + L_{mem} + L_{ret}$.

%% file: experiment.tex
\section{Experimental Results}
\label{sec:result}

\subsection{Data collection}
\label{sec:collection}
We used an {\it Intel RealSense 415} camera to record several  demonstrations of a human subject performing three tasks. The first task consists in inserting a paint brush into a bucket, then moving it to a painting surface and painting a virtual straight line on the surface. The poses of the brush, bucket and painting surface are all tracked in real-time using the technique explained in~\ref{sec:tracking}. 
The second task consists in picking up various blocks and stacking them on top of each other to form a predefined desired pattern. The third task is similar to the second one, with the only difference being the desired stacking pattern. Additionally, we use the PyBullet physics engine to simulate a {\it Kuka} robotic arm and collect data regarding a fourth task. The fourth task consists in moving a wrench that is attached to the end-effector to four precise locations on a wheel, sequentially, rotating the wrench at each location to remove the lug-nuts, then moving the wrench to the wheel's center before finally pulling it. 


\begin{figure}[h]
    \vspace{-.1in}
    \begin{center}
     \includegraphics[width=0.12\textwidth]{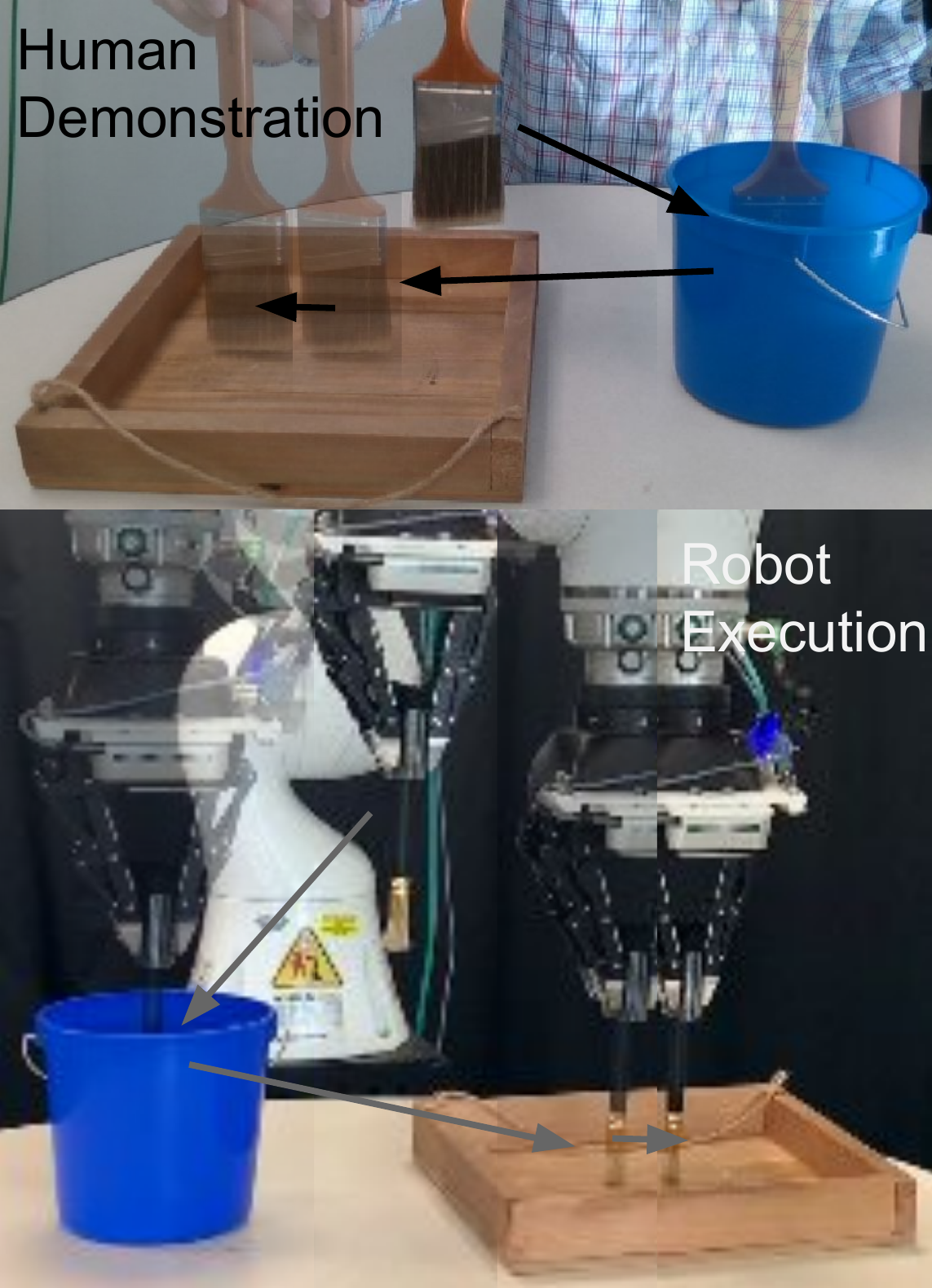}
     \includegraphics[width=0.093\textwidth]{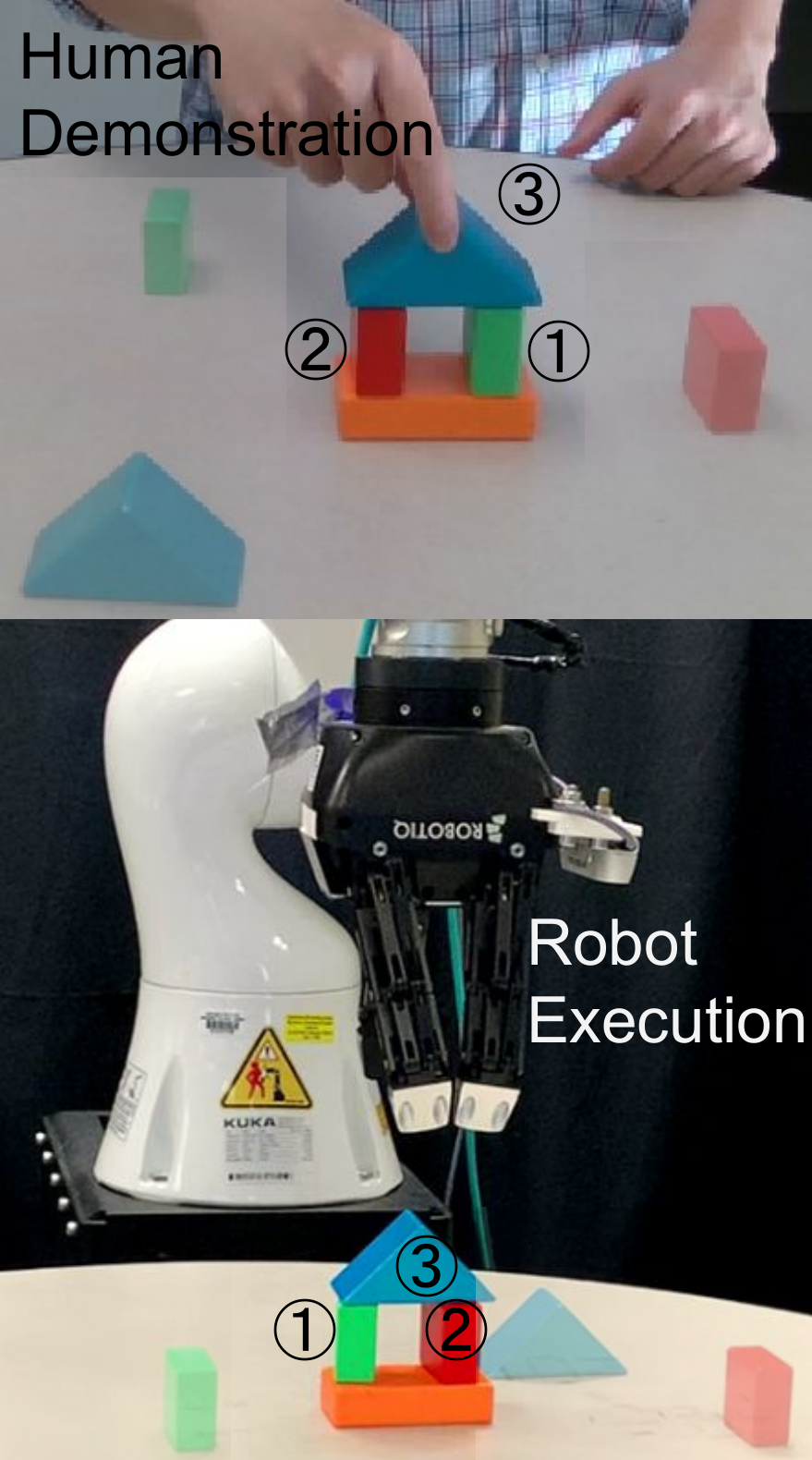}
     \includegraphics[width=0.095\textwidth]{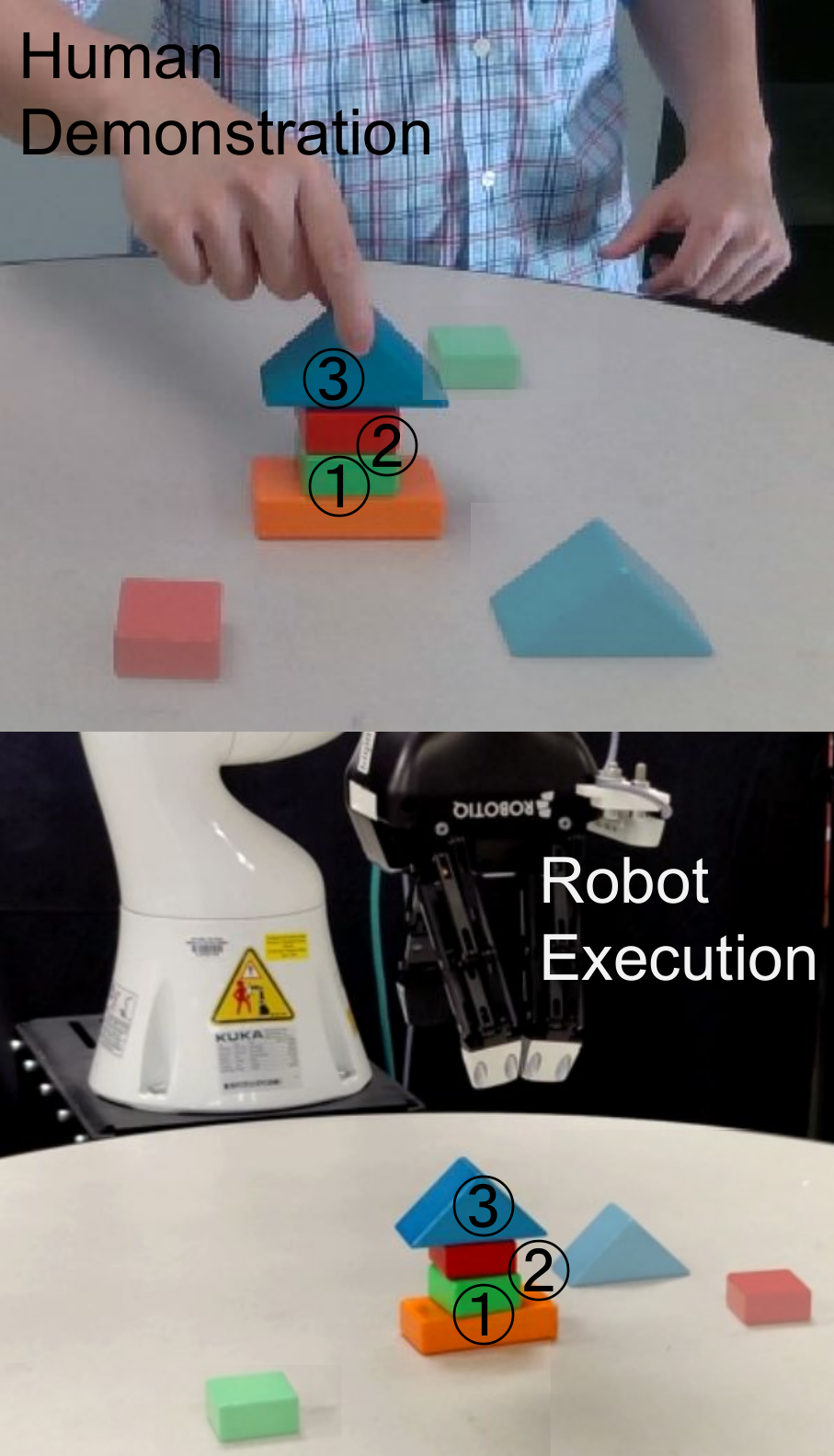}
     \includegraphics[width=0.13\textwidth]{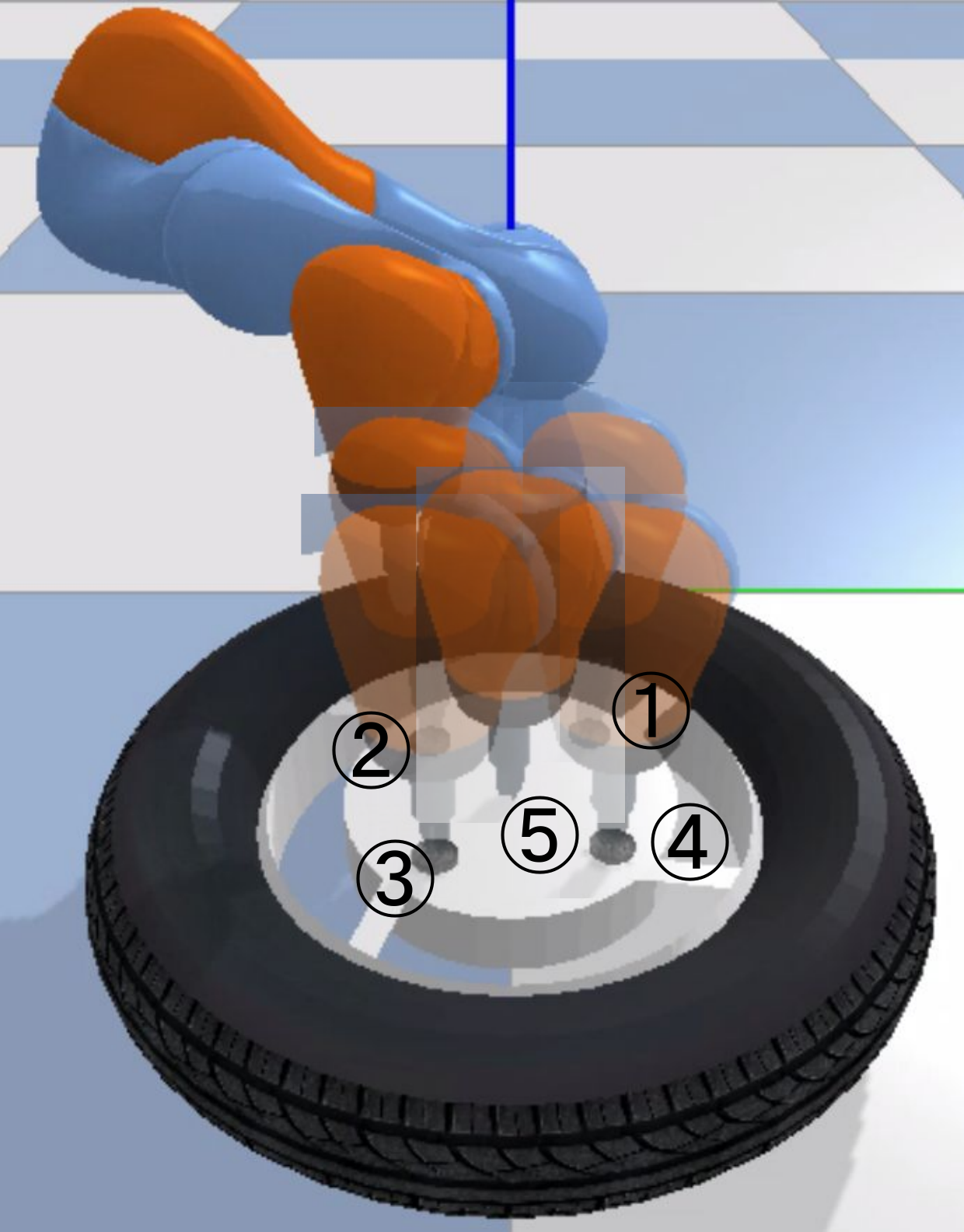}
    \end{center}
    \caption{Tasks considered in the experiments: painting (left),  stacking (middle), and tire removal (right).}
    \vspace{-.2in}
\label{fig:setups}
\end{figure}

\begin{figure}
     \begin{subfigure}{0.24\textwidth}
     \includegraphics[width=\textwidth]{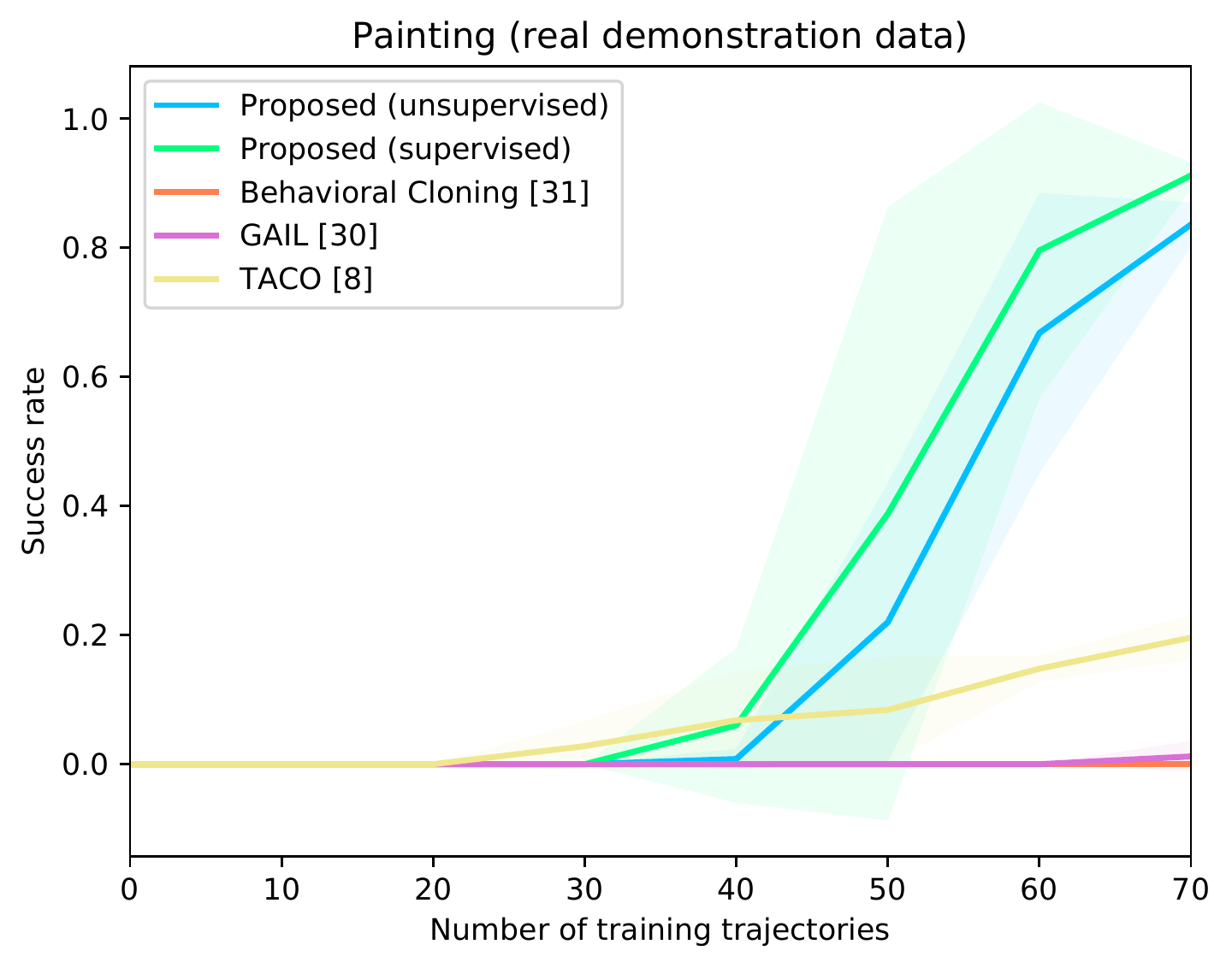}
     \end{subfigure}
     \begin{subfigure}{0.24\textwidth}
     \includegraphics[width=\textwidth]{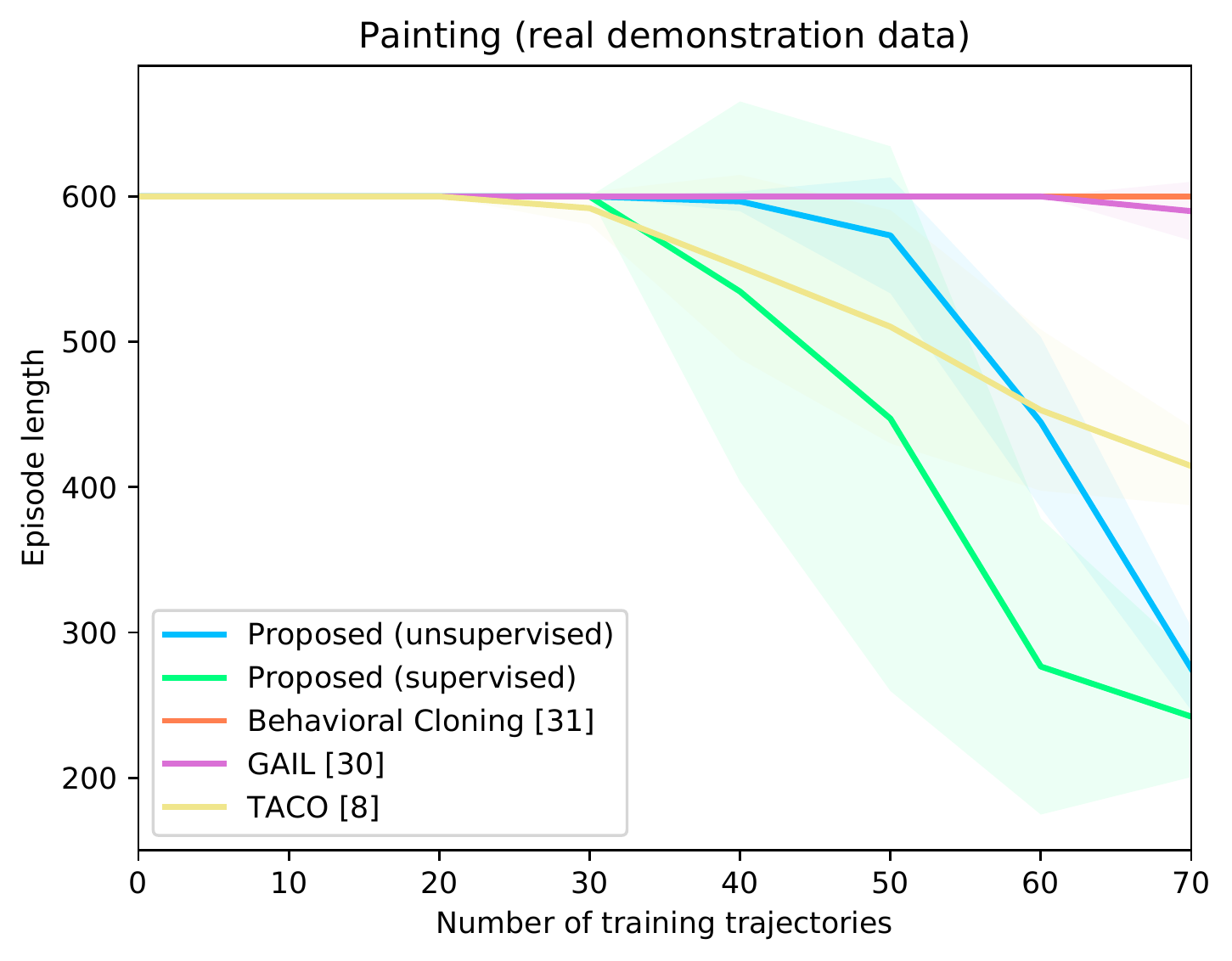}
     \end{subfigure}
     \begin{subfigure}{0.24\textwidth}
     \includegraphics[width=\textwidth]{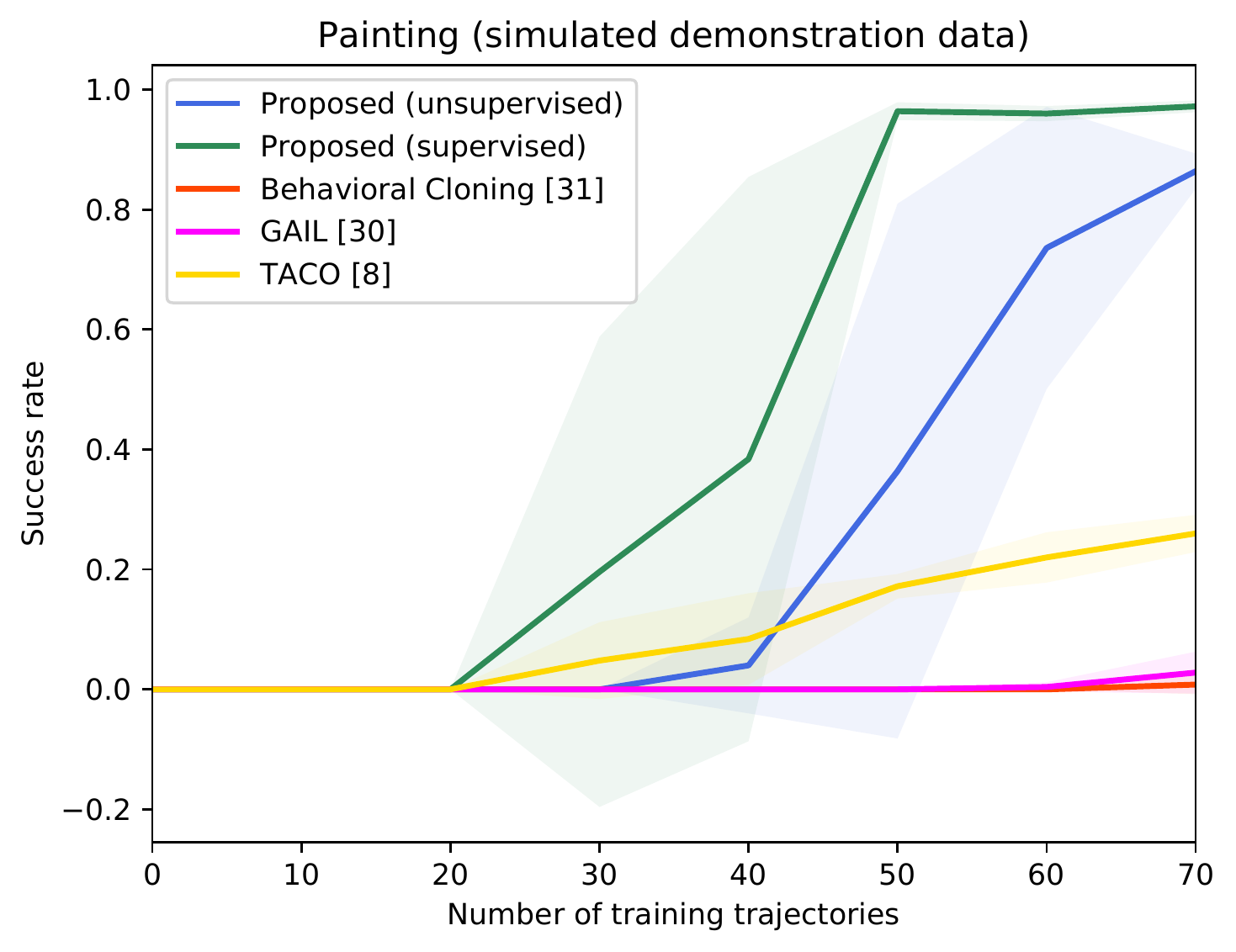}
     \end{subfigure}
     \begin{subfigure}{0.24\textwidth}
     \includegraphics[width=\textwidth]{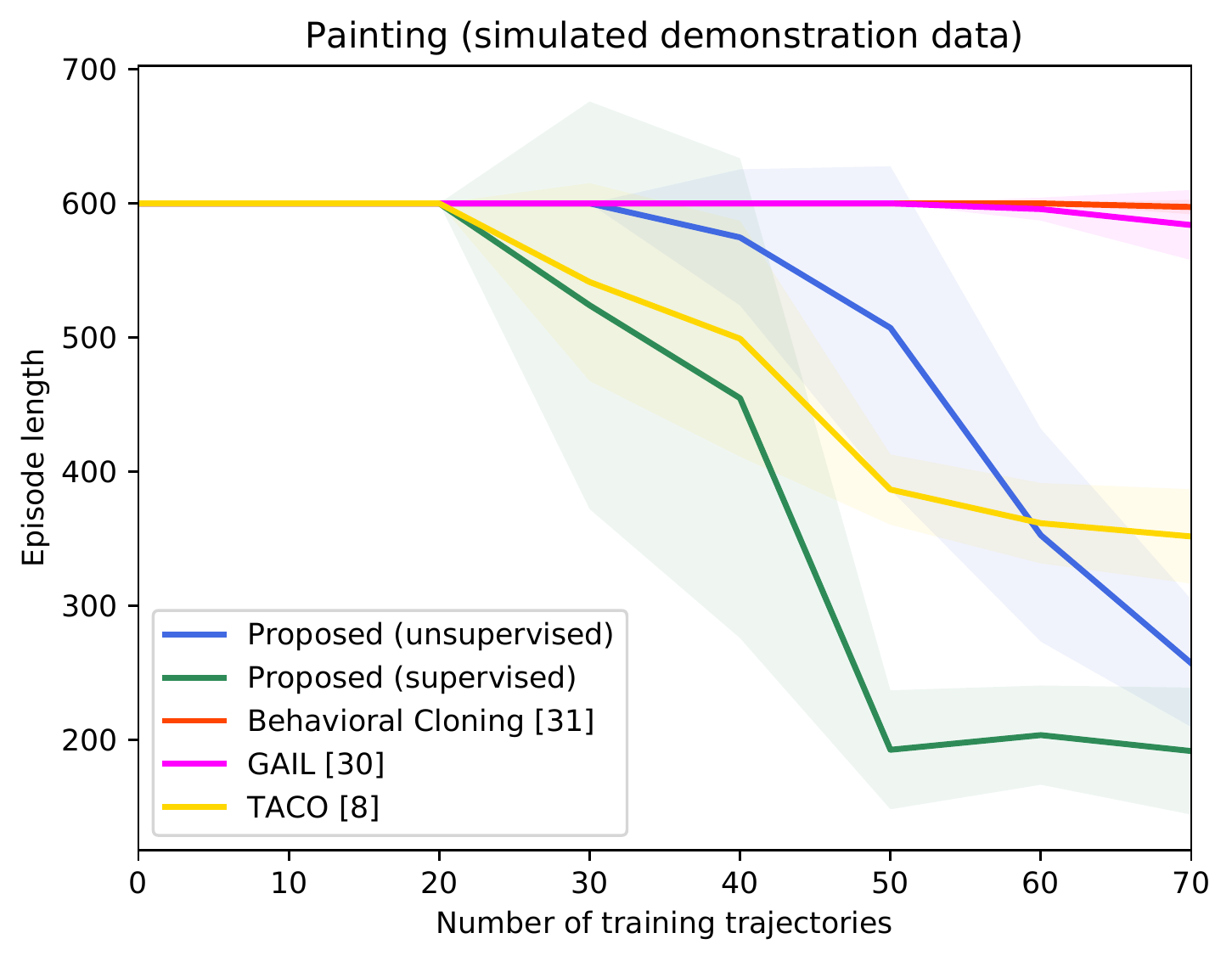}
     \end{subfigure}
     \begin{subfigure}{0.24\textwidth}
     \includegraphics[width=\textwidth]{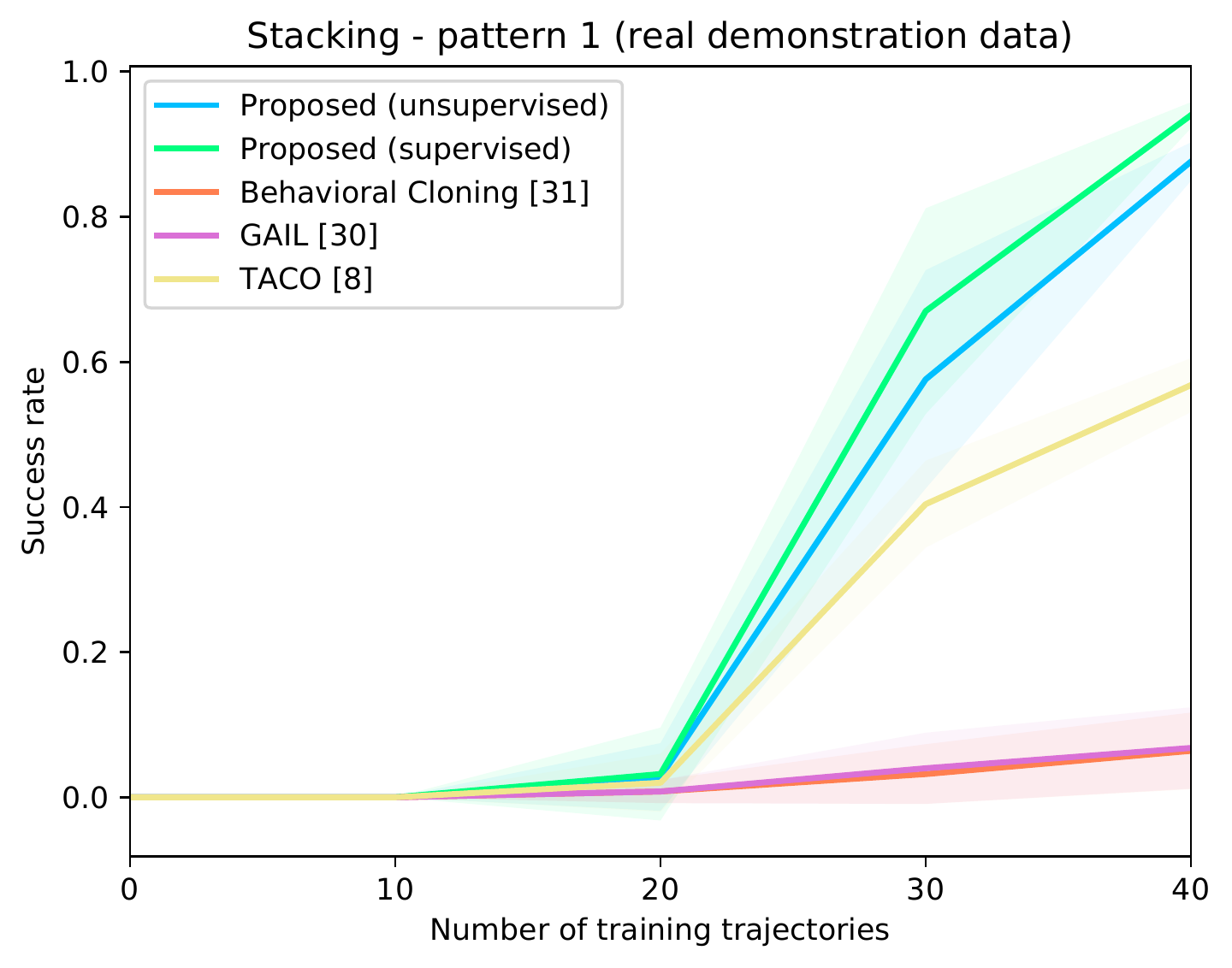}
     \end{subfigure}
     \begin{subfigure}{0.24\textwidth}
     \includegraphics[width=\textwidth]{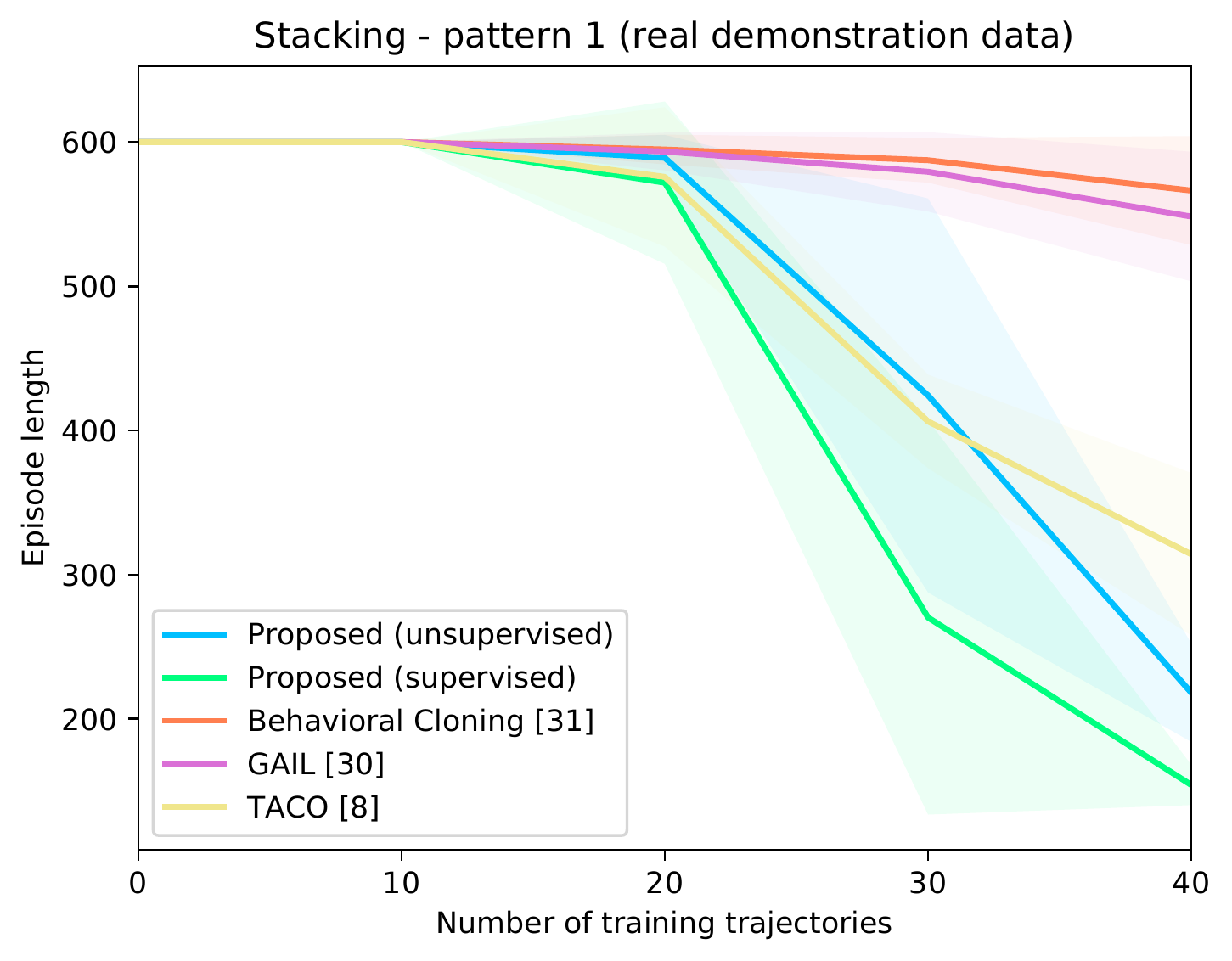}
     \end{subfigure}
     \begin{subfigure}{0.24\textwidth}
     \includegraphics[width=\textwidth]{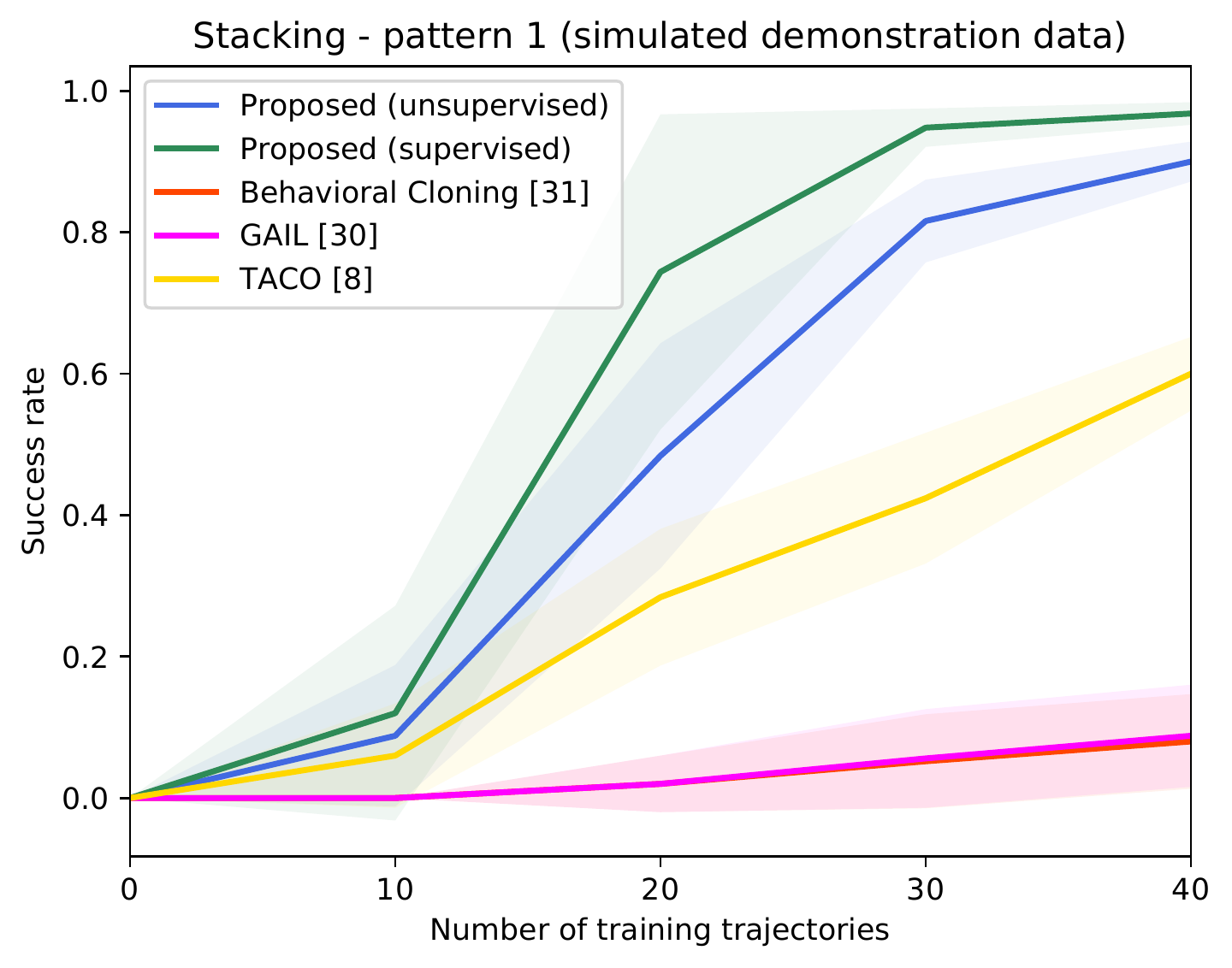}
     \end{subfigure}
     \begin{subfigure}{0.24\textwidth}
     \includegraphics[width=\textwidth]{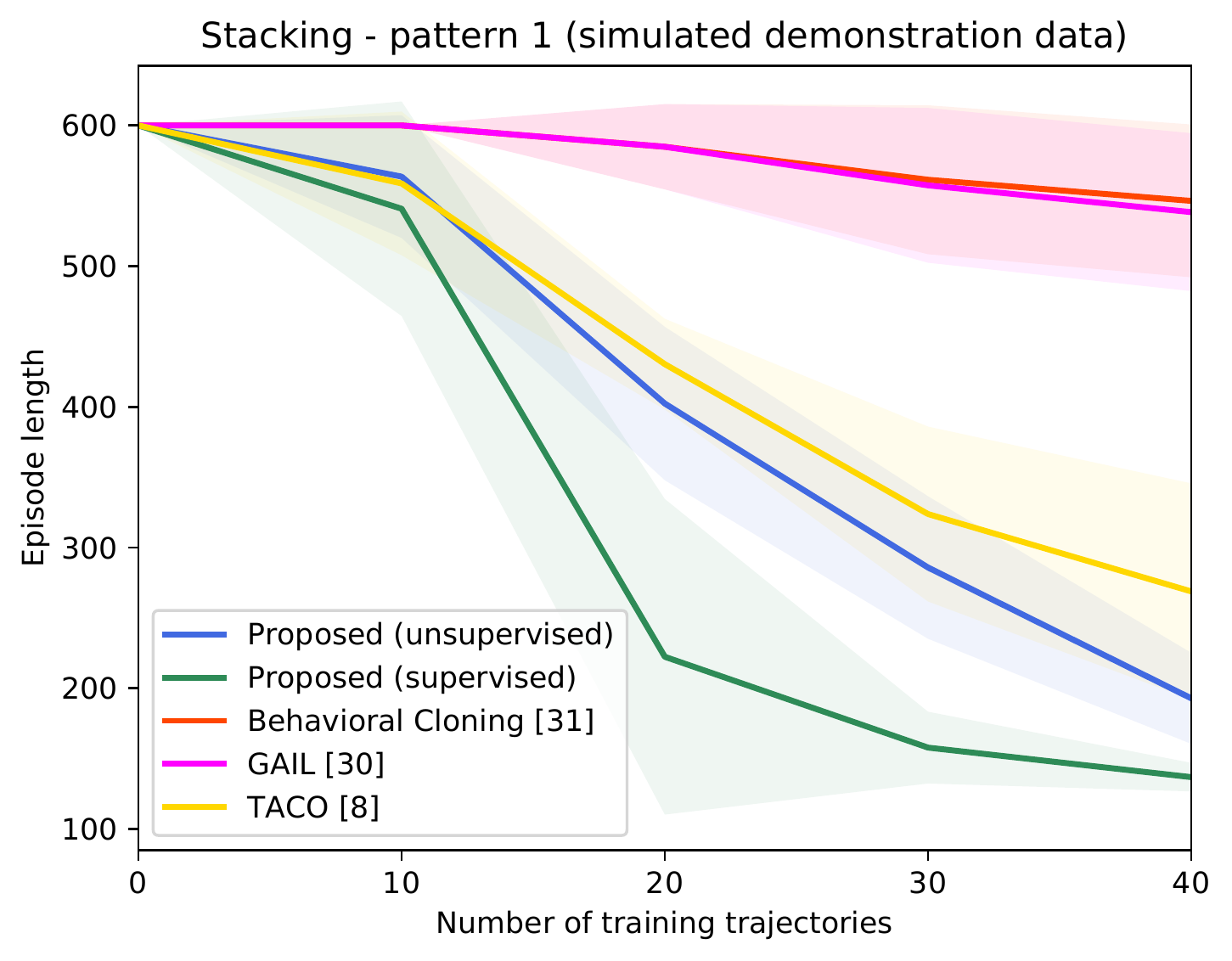}
     \end{subfigure}
     \begin{subfigure}{0.24\textwidth}
     \includegraphics[width=\textwidth]{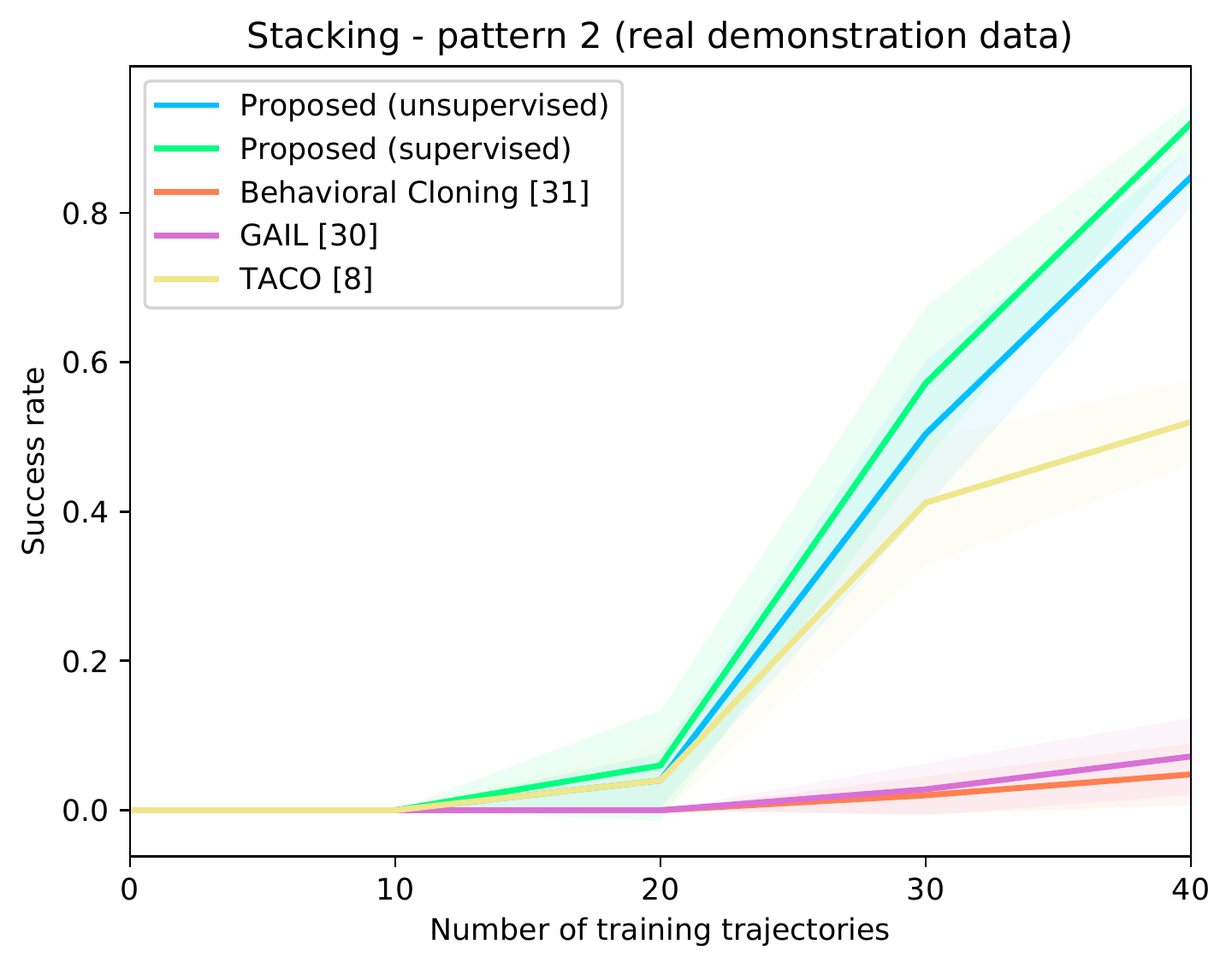}
     \end{subfigure}
     \begin{subfigure}{0.24\textwidth}
     \includegraphics[width=\textwidth]{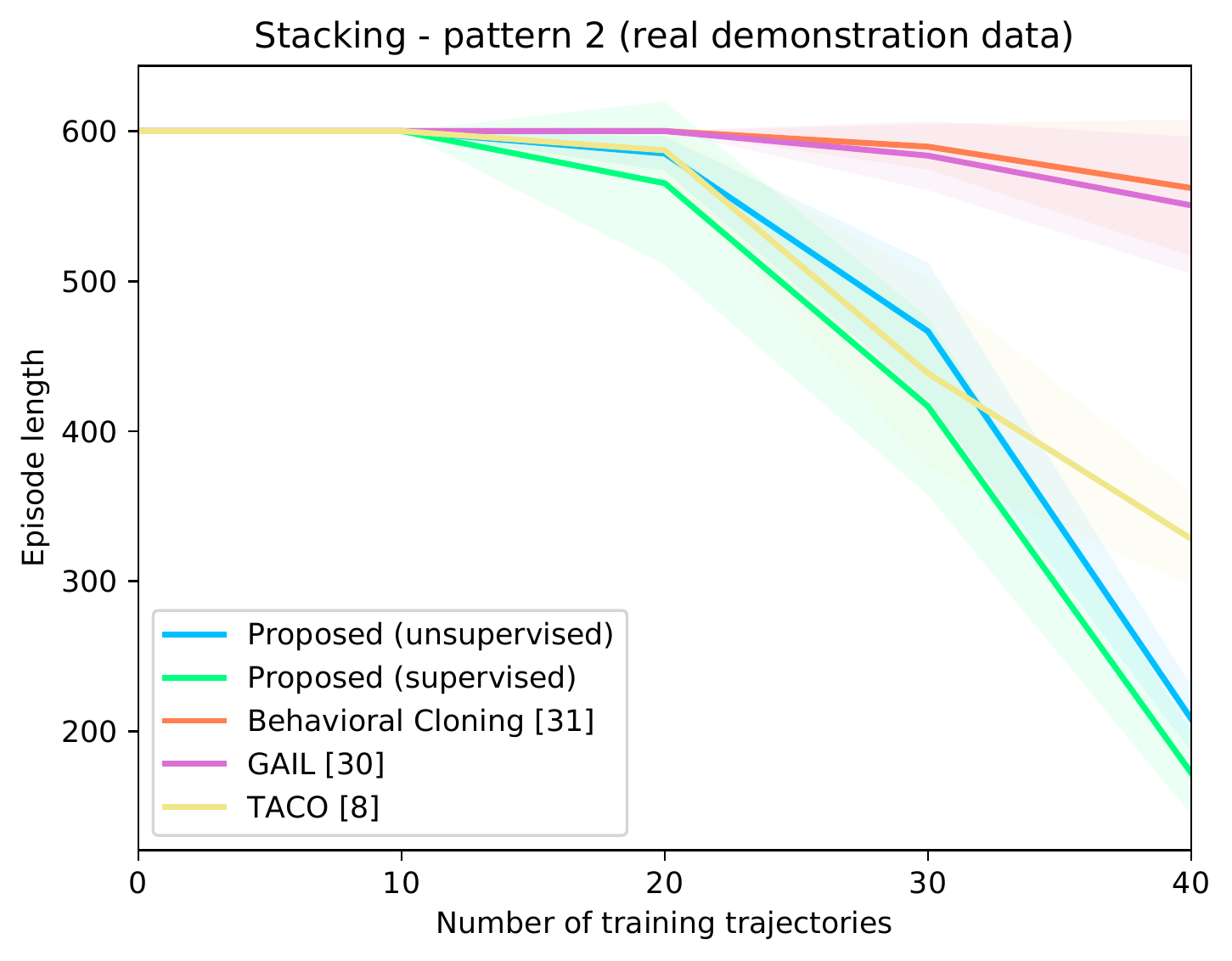}
     \end{subfigure}
     \begin{subfigure}{0.24\textwidth}
     \includegraphics[width=\textwidth]{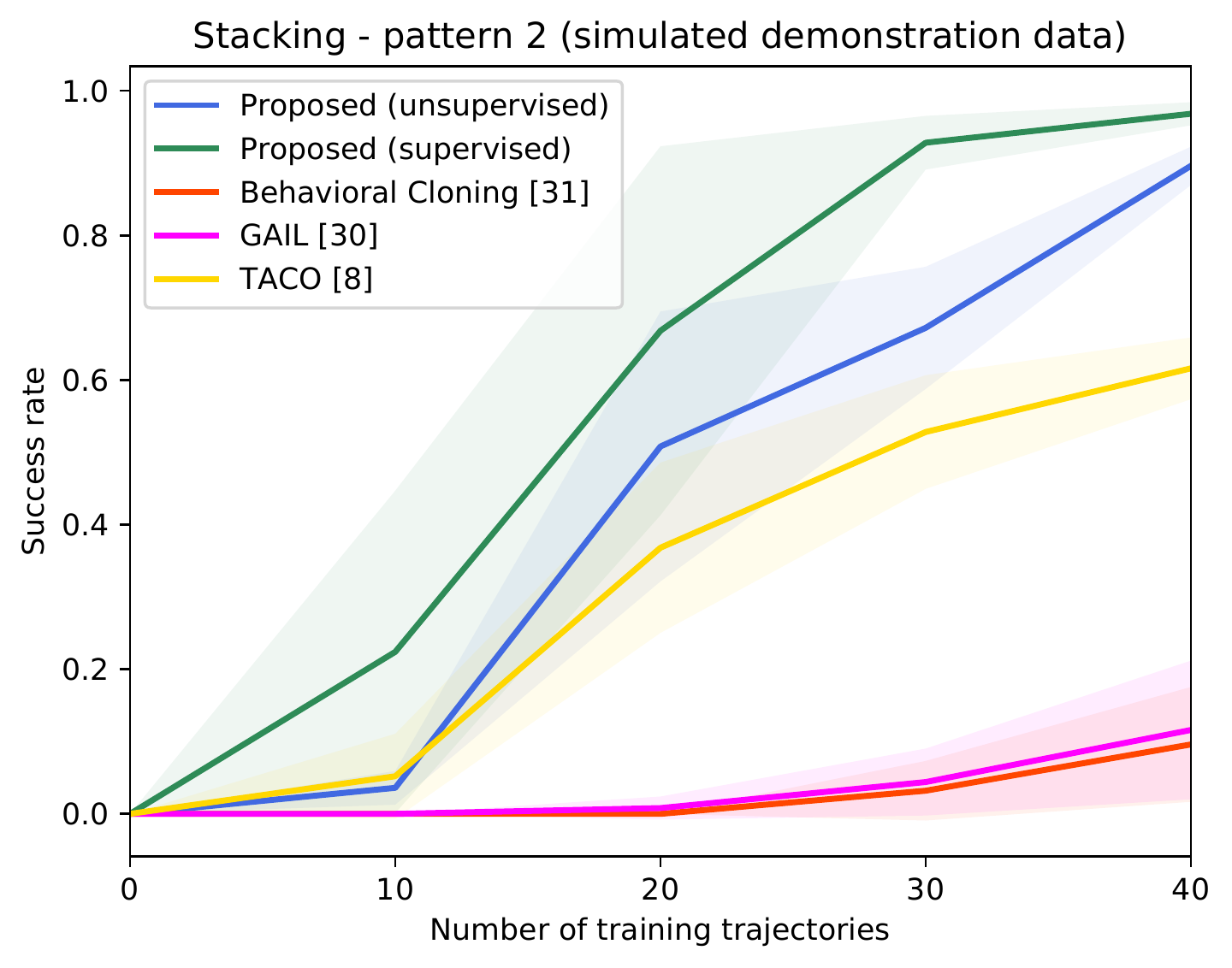}
     \end{subfigure}
     \begin{subfigure}{0.24\textwidth}
     \includegraphics[width=\textwidth]{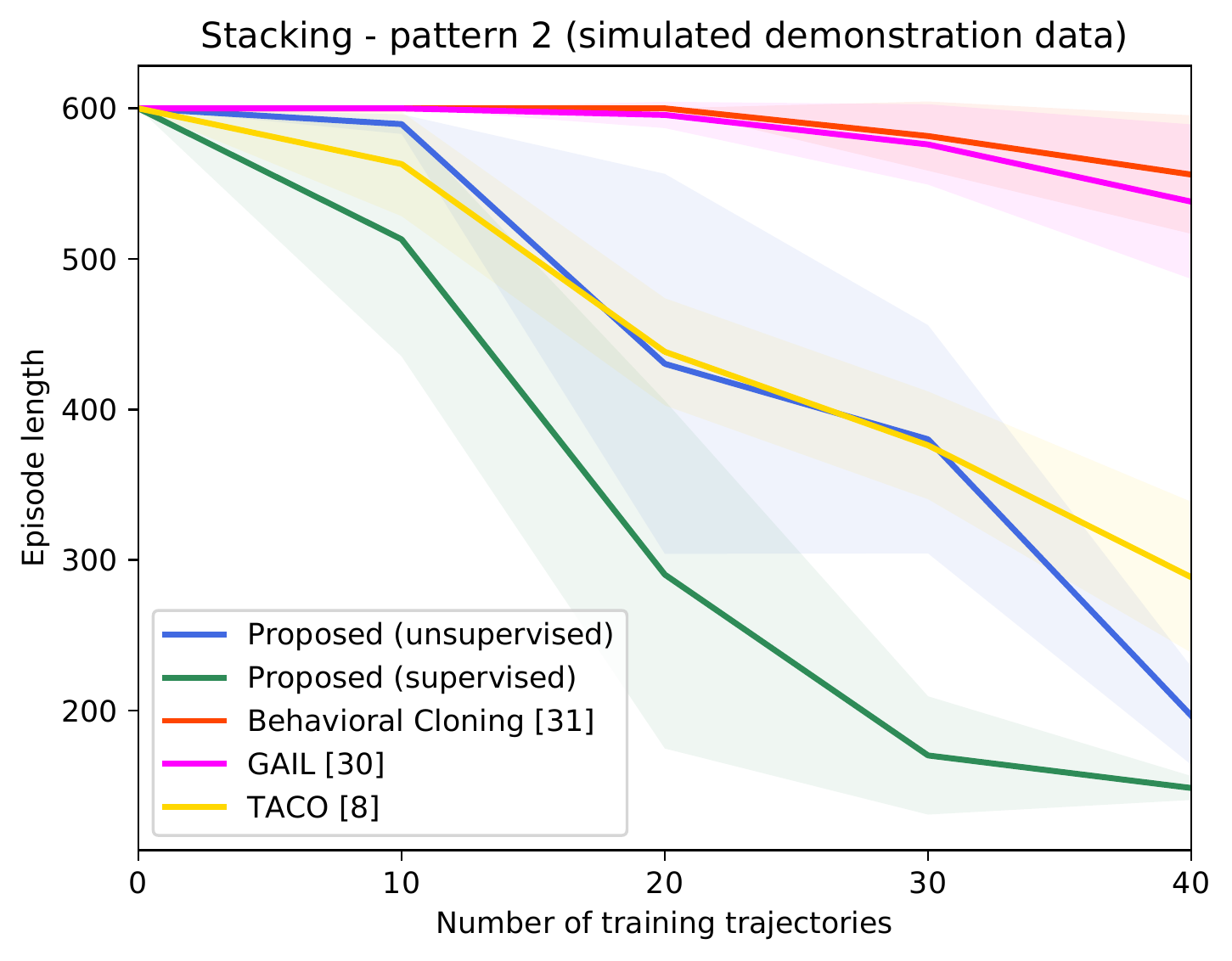}
     \end{subfigure}
     \begin{subfigure}{0.24\textwidth}
     \includegraphics[width=\textwidth]{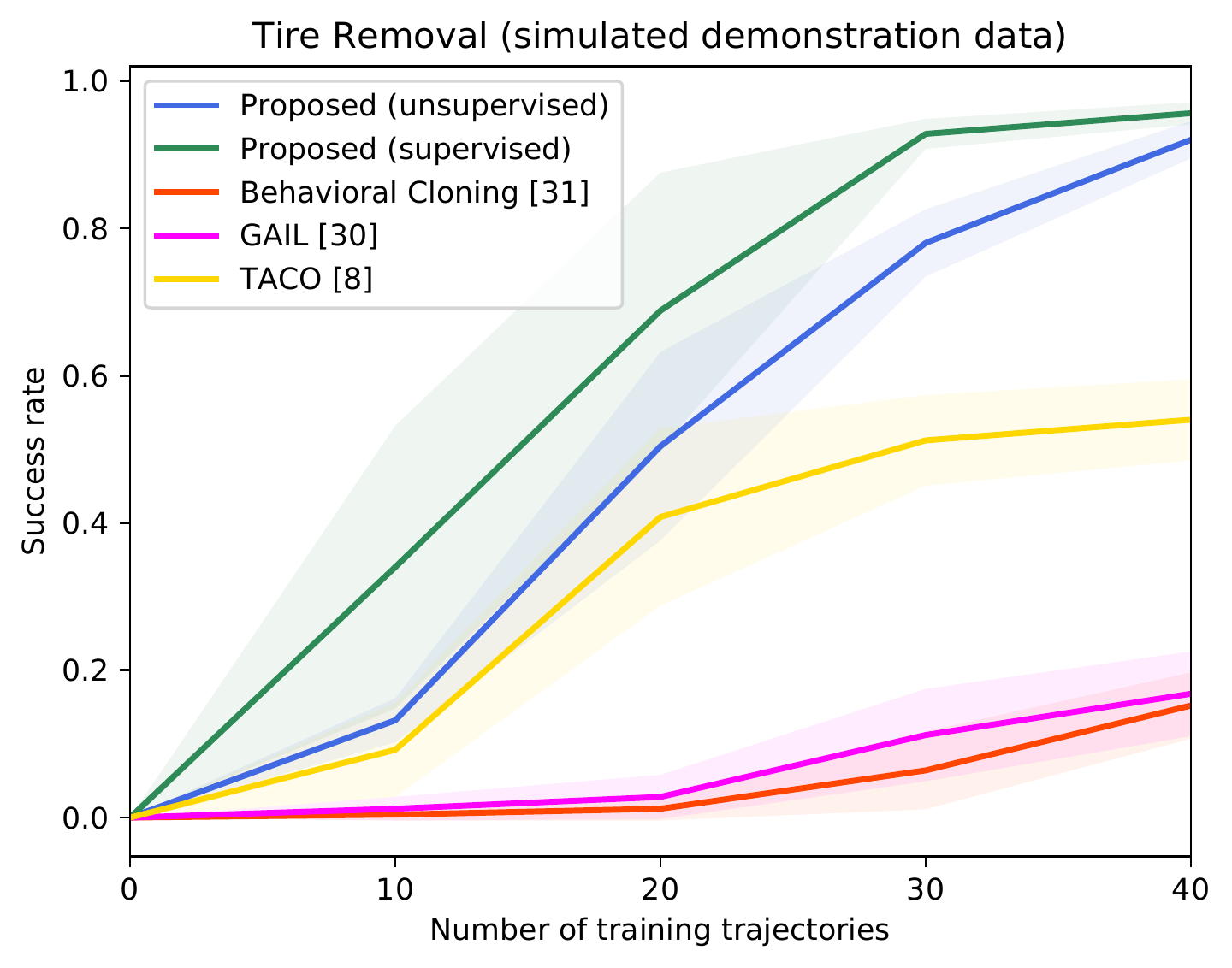}
     \end{subfigure}
     \begin{subfigure}{0.24\textwidth}
     \includegraphics[width=\textwidth]{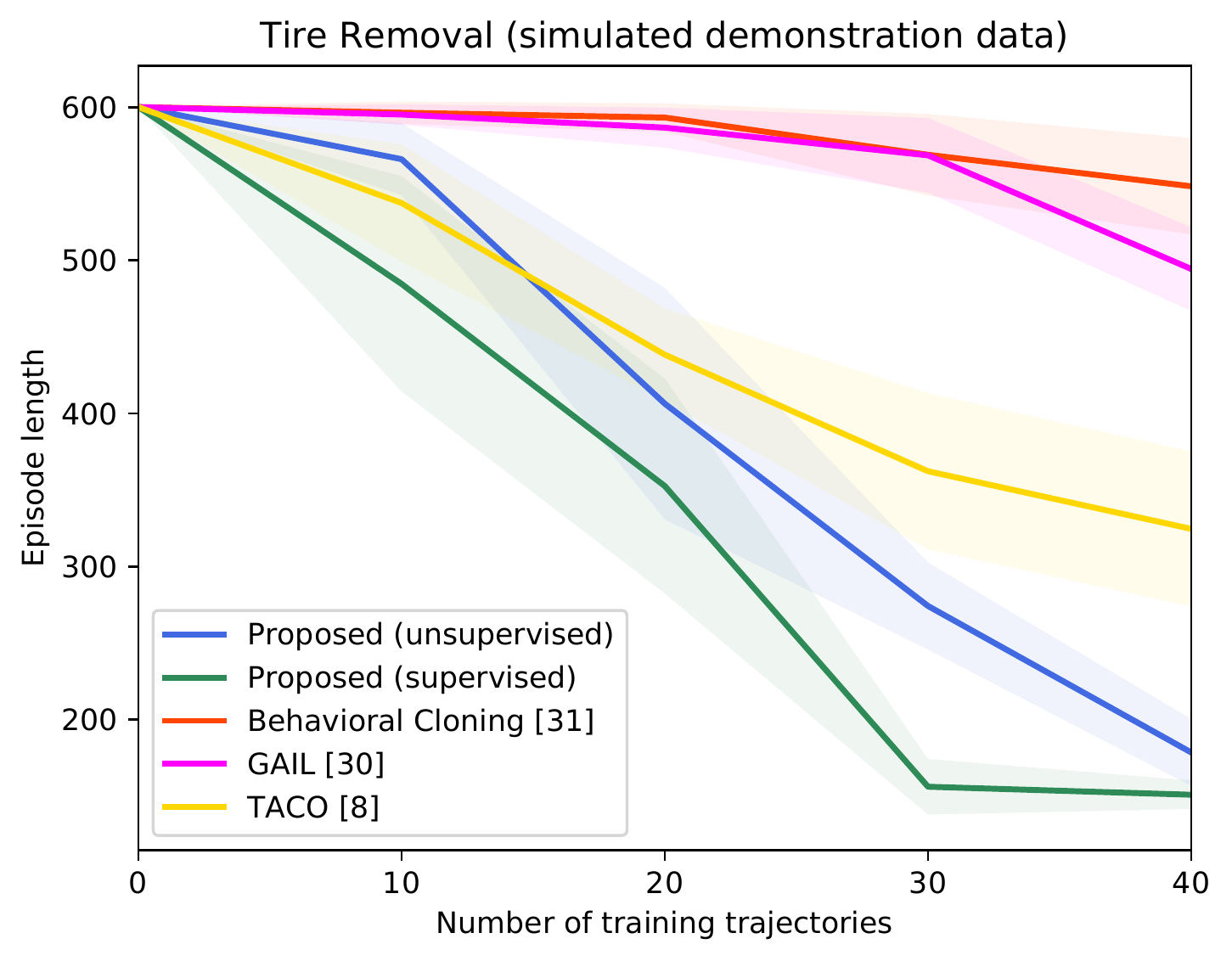}
     \end{subfigure}
     \caption{Average success rate (left) and episode length (right) as a function of the number of training trajectories}
\label{fig:success_rate}
\end{figure}

\subsection{Object Pose Parsing from Demonstration Video}
\label{sec:tracking}

In each demonstration video, 6D poses $\xi \in SE(3)$ and semantic labels of all relevant objects are estimated in real-time and used to create observations $(z_t)_{t=1}^{H}$ as explained in Section~\ref{sec:model}. Concretely, a scene-level multi-object pose estimator~\cite{mitash2020scene} is leveraged to compute globally the relevant objects' 6D poses in the first frame. It starts with a pose sampling process and performs Integer Linear Programming to ensure physical consistency by checking collisions between any two objects as well as collisions between any object and the table. Next, the poses computed from the first frame are used to initialize the se(3)-TrackNet~\cite{wen2020se}, which returns a 6D pose for each object in every frame of the video. The 6D tracker requires access to the objects' CAD models. For the painting task, the brush and the bucket are 3D-scanned as in \cite{morgan2021vision}, while for all the other objects, models are obtained from CAD designs derived from geometric primitives. During inference on the demonstration videos, the tracker operates in $90 Hz$, resulting in an average processing time of $13.3s$ for a 1 min demonstration video. 
The entire video parsing process is fully automated, and did not require any human input beyond providing CAD models of the objects offline to the se(3)-TrackNet in order to learn to track them.

\subsection{Training and architecture details}
The high-level, intermediate-level, and low-level policies are all neural networks. In the high-level policy, the progress vector $g_t$ is embedded by a fully connected layer with $64$ units followed by a ReLU layer. An LSTM layer with $32$ units is used to encode history $h_t$. Observation $z_t$ is concatenated with the LSTM units and the embedded progress vector and fed as an input to two hidden layers with $64$ units followed by a ReLU layer. From the last hidden layer, target and tool objects are predicted by a fully connected layer and a softmax. The intermediate-level policy network consists of two hidden layers with $64$ units. The low-level policy concatenates into a vector four inputs: 6D pose of the target in the frame of the tool object, way-point, and the semantic labels of the target and tool objects. After two hidden layers of $64$ units and ReLU, the low-level policy outputs a Gaussian action distribution. The number of training iterations for all tasks is $20,000$, the batch size is $2,048$ steps, the learning rate is $1e-4$, and the optimizer is {\it Adam}. The hyper-parameters used in Section~\ref{sec:initialization} are set as $\gamma = 0.95$, $\alpha=100$, and $\beta=0.95$.

The proposed method is compared to three other techniques. Generative Adversarial Imitation Learning ({\bf GAIL})~\cite{ho2016generative}, Learning Task Decomposition via Temporal Alignment for Control ({\bf TACO})~\cite{shiarlis2018taco}, and {\bf Behavioral Cloning}~\cite{pomerleau1989alvinn} where we train the policy network of~\cite{ho2016generative} directly to maximize the likelihood of the demonstrations without learning rewards. We also compare to a supervised variant of our proposed technique where we manually provide ground-truth tool and target objects and way-points for each frame. The supervised variant provides an upper bound on the performance of our unsupervised algorithm. Note also that TACO~\cite{shiarlis2018taco} requires providing manual {\it sketches} of the sub-tasks, whereas our algorithm is fully unsupervised. 


\subsection{Evaluation}
Except for the tire removal, all tasks are evaluated using real demonstrations and a real {\it Kuka LBR} robot equipped with a {\it Robotiq} hand. The policies learned from real demonstrations are also tested extensively in the PyBullet simulator before testing them on the real robot.

A painting task is successfully accomplished if the brush is moved into a specific region of a radius of $3cm$ inside the paint bucket, the brush is then moved to a plane that is $10cm$ on top of the painting surface, and finally the brush draws a virtual straight line of $5cm$ at least on that plane. A tire removal task is successfully accomplished if the robot removes all bolts by rotating its end-effector on top of each bolt (with a toleance of $5mm$) with at least $30^\circ$ counter-clockwise, and then moves to the center of the wheel. A stacking task is successful if the centers of all the objects in their final configuration are within $0.5cm$ of the corresponding desired locations.

\begin{table}
\begin{center}
\begin{tabular}{ | c | c | c | c |}
  \hline
   & Painting & Stacking  1 & Stacking 2 \\ 
  \hline
 Proposed (unsupervised) & $5/5$ & $5/5$ & $5/5$ \\  
  \hline
 Behavioral Cloning \cite{pomerleau1989alvinn} & $0/5$ & $0/5$ & $0/5$ \\
  \hline
 GAIL \cite{ho2016generative} & $0/5$ & $0/5$ & $0/5$ \\
  \hline
 TACO \cite{shiarlis2018taco} & $1/5$ & $1/5$ & $2/5$ \\
  \hline
\end{tabular}
\end{center}
\caption{Success rates on the real Kuka robot}
\label{table:real_eval}
\vspace{-0.5cm}
\end{table}

Figure~\ref{fig:success_rate} shows the success rates of the compared methods for the four tasks, as well as the length of the generated trajectories while solving these tasks in simulation, as a function of the number of demonstration trajectories collected as explained in Section~\ref{sec:collection}. The results are averaged over $5$ independent runs, each run contains $50$ test episodes that start with random layouts of the objects. Table~\ref{table:real_eval} shows the success rates of the compared methods on the real Kuka robot, using the same demonstration trajectories that were used to generate Figure~\ref{fig:success_rate} ($70$ trajectories for painting and $40$ for each of the remaining tasks).
These results show clearly that the proposed approach outperforms the compared alternatives in terms of success rates and solves the four tasks with a smaller number of actions. The performance of our unsupervised approach is also close to that of the supervised variant. The proposed approach outperforms TACO despite the fact that TACO requires a form of supervision in its training. We also note that both our approach and TACO outperform GAIL and the behavioral cloning techniques, which clearly indicates the data-efficiency of compositional and hierarchical methods.
Videos and supplementary material can be found at \textcolor{red}{\bf \url{https://tinyurl.com/2zrp2rzm}}.

%% file: main.bbl
\begin{thebibliography}{10}
\providecommand{\url}[1]{#1}
\csname url@rmstyle\endcsname
\providecommand{\newblock}{\relax}
\providecommand{\bibinfo}[2]{#2}
\providecommand\BIBentrySTDinterwordspacing{\spaceskip=0pt\relax}
\providecommand\BIBentryALTinterwordstretchfactor{4}
\providecommand\BIBentryALTinterwordspacing{\spaceskip=\fontdimen2\font plus
\BIBentryALTinterwordstretchfactor\fontdimen3\font minus
  \fontdimen4\font\relax}
\providecommand\BIBforeignlanguage[2]{{%
\expandafter\ifx\csname l@#1\endcsname\relax
\typeout{** WARNING: IEEEtran.bst: No hyphenation pattern has been}%
\typeout{** loaded for the language `#1'. Using the pattern for}%
\typeout{** the default language instead.}%
\else
\language=\csname l@#1\endcsname
\fi
#2}}

\bibitem{10.5555/1795482}
S.~Calinon, \emph{Robot Programming by Demonstration}, 1st~ed.\hskip 1em plus
  0.5em minus 0.4em\relax USA: CRC Press, Inc., 2009.

\bibitem{10.5555/3235188}
T.~Osa, J.~Pajarinen, and G.~Neumann, \emph{An Algorithmic Perspective on
  Imitation Learning}.\hskip 1em plus 0.5em minus 0.4em\relax Hanover, MA, USA:
  Now Publishers Inc., 2018.

\bibitem{Kroemer-2016-954}
\BIBentryALTinterwordspacing
O.~Kroemer and G.~S. Sukhatme, ``Learning spatial preconditions of manipulation
  skills using random forests,'' in \emph{Proceedings of the IEEE-RAS
  International Conference on Humanoid Robotics}, 2016. [Online]. Available:
  \url{http://robotics.usc.edu/publications/954/}
\BIBentrySTDinterwordspacing

\bibitem{Wang-2019-112298}
A.~S. Wang and O.~Kroemer, ``Learning robust manipulation strategies with
  multimodal state transition models and recovery heuristics,'' in
  \emph{Proceedings of (ICRA) International Conference on Robotics and
  Automation}, May 2019, pp. 1309 -- 1315.

\bibitem{DBLP:journals/ijrr/WangGKL21}
\BIBentryALTinterwordspacing
Z.~Wang, C.~R. Garrett, L.~P. Kaelbling, and T.~Lozano{-}P{\'{e}}rez,
  ``Learning compositional models of robot skills for task and motion
  planning,'' \emph{Int. J. Robotics Res.}, vol.~40, no. 6-7, 2021. [Online].
  Available: \url{https://doi.org/10.1177/02783649211004615}
\BIBentrySTDinterwordspacing

\bibitem{Su-2016-112221}
Z.~Su, O.~Kroemer, G.~E. Loeb, G.~S. Sukhatme, and S.~Schaal, ``Learning to
  switch between sensorimotor primitives using multimodal haptic signals,'' in
  \emph{Proceedings of International Conference on Simulation of Adaptive
  Behavior (SAB '16): From Animals to Animats 14}, August 2016, pp. 170 -- 182.

\bibitem{le2018hierarchical}
H.~Le, N.~Jiang, A.~Agarwal, M.~Dud{\'\i}k, Y.~Yue, and H.~Daum{\'e},
  ``Hierarchical imitation and reinforcement learning,'' in \emph{International
  conference on machine learning}.\hskip 1em plus 0.5em minus 0.4em\relax PMLR,
  2018, pp. 2917--2926.

\bibitem{shiarlis2018taco}
K.~Shiarlis, M.~Wulfmeier, S.~Salter, S.~Whiteson, and I.~Posner, ``Taco:
  Learning task decomposition via temporal alignment for control,'' in
  \emph{International Conference on Machine Learning}.\hskip 1em plus 0.5em
  minus 0.4em\relax PMLR, 2018, pp. 4654--4663.

\bibitem{NIPS2009_e0cf1f47}
G.~Konidaris and A.~Barto, ``Skill discovery in continuous reinforcement
  learning domains using skill chaining,'' in \emph{Advances in Neural
  Information Processing Systems}, Y.~Bengio, D.~Schuurmans, J.~Lafferty,
  C.~Williams, and A.~Culotta, Eds., vol.~22.\hskip 1em plus 0.5em minus
  0.4em\relax Curran Associates, Inc., 2009.

\bibitem{DBLP:conf/ijcai/ToussaintAST19}
\BIBentryALTinterwordspacing
M.~Toussaint, K.~R. Allen, K.~A. Smith, and J.~B. Tenenbaum, ``Differentiable
  physics and stable modes for tool-use and manipulation planning,'' in
  \emph{Proceedings of the Twenty-Eighth International Joint Conference on
  Artificial Intelligence, {IJCAI} 2019, Macao, China, August 10-16, 2019},
  S.~Kraus, Ed.\hskip 1em plus 0.5em minus 0.4em\relax ijcai.org, 2019, pp.
  6231--6235. [Online]. Available:
  \url{https://doi.org/10.24963/ijcai.2019/869}
\BIBentrySTDinterwordspacing

\bibitem{Kaelbling93learningto}
L.~P. Kaelbling, ``Learning to achieve goals,'' in \emph{IN PROC. OF
  IJCAI-93}.\hskip 1em plus 0.5em minus 0.4em\relax Morgan Kaufmann, 1993, pp.
  1094--1098.

\bibitem{10.5555/2908515.2908520}
L.~P. Kaelbling and T.~Lozano-P\'{e}rez, ``Hierarchical task and motion
  planning in the now,'' in \emph{Proceedings of the 1st AAAI Conference on
  Bridging the Gap Between Task and Motion Planning}, ser. AAAIWS'10-01.\hskip
  1em plus 0.5em minus 0.4em\relax AAAI Press, 2010, pp. 33--42.

\bibitem{garrett2020integrated}
C.~R. Garrett, R.~Chitnis, R.~Holladay, B.~Kim, T.~Silver, L.~P. Kaelbling, and
  T.~Lozano-Pérez, ``Integrated task and motion planning,'' 2020.

\bibitem{pmlr-v80-icarte18a}
R.~T. Icarte, T.~Klassen, R.~Valenzano, and S.~McIlraith, ``Using reward
  machines for high-level task specification and decomposition in reinforcement
  learning,'' ser. Proceedings of Machine Learning Research, J.~Dy and
  A.~Krause, Eds., vol.~80.\hskip 1em plus 0.5em minus 0.4em\relax
  StockholmsmÃ€ssan, Stockholm Sweden: PMLR, 10--15 Jul 2018, pp.
  2107--2116.

\bibitem{toro2019learning}
R.~Toro~Icarte, E.~Waldie, T.~Klassen, R.~Valenzano, M.~Castro, and
  S.~McIlraith, ``Learning reward machines for partially observable
  reinforcement learning,'' \emph{Advances in Neural Information Processing
  Systems}, vol.~32, pp. 15\,523--15\,534, 2019.

\bibitem{camacho2019ltl}
A.~Camacho, R.~T. Icarte, T.~Q. Klassen, R.~A. Valenzano, and S.~A. McIlraith,
  ``Ltl and beyond: Formal languages for reward function specification in
  reinforcement learning.''

\bibitem{wen2020se}
B.~Wen, C.~Mitash, B.~Ren, and K.~E. Bekris, ``se (3)-tracknet: Data-driven 6d
  pose tracking by calibrating image residuals in synthetic domains,'' in
  \emph{2020 IEEE/RSJ International Conference on Intelligent Robots and
  Systems (IROS)}.\hskip 1em plus 0.5em minus 0.4em\relax IEEE, pp.
  10\,367--10\,373.

\bibitem{pmlr-v87-kalashnikov18a}
D.~Kalashnikov, A.~Irpan, P.~Pastor, J.~Ibarz, A.~Herzog, E.~Jang, D.~Quillen,
  E.~Holly, M.~Kalakrishnan, V.~Vanhoucke, and S.~Levine, ``Scalable deep
  reinforcement learning for vision-based robotic manipulation,'' ser.
  Proceedings of Machine Learning Research, A.~Billard, A.~Dragan, J.~Peters,
  and J.~Morimoto, Eds., vol.~87.\hskip 1em plus 0.5em minus 0.4em\relax PMLR,
  29--31 Oct 2018, pp. 651--673.

\bibitem{fox2018parametrized}
\BIBentryALTinterwordspacing
R.~Fox, R.~Shin, S.~Krishnan, K.~Goldberg, D.~Song, and I.~Stoica,
  ``Parametrized hierarchical procedures for neural programming,'' in
  \emph{International Conference on Learning Representations}, 2018. [Online].
  Available: \url{https://openreview.net/forum?id=rJl63fZRb}
\BIBentrySTDinterwordspacing

\bibitem{journals/corr/abs-1710-01813}
\BIBentryALTinterwordspacing
D.~Xu, S.~Nair, Y.~Zhu, J.~Gao, A.~Garg, L.~Fei-Fei, and S.~Savarese, ``Neural
  task programming: Learning to generalize across hierarchical tasks.''
  \emph{CoRR}, vol. abs/1710.01813, 2017. [Online]. Available:
  \url{http://dblp.uni-trier.de/db/journals/corr/corr1710.html#abs-1710-01813}
\BIBentrySTDinterwordspacing

\bibitem{DBLP:journals/corr/abs-1807-03480}
\BIBentryALTinterwordspacing
D.~Huang, S.~Nair, D.~Xu, Y.~Zhu, A.~Garg, L.~Fei{-}Fei, S.~Savarese, and J.~C.
  Niebles, ``Neural task graphs: Generalizing to unseen tasks from a single
  video demonstration,'' \emph{CoRR}, vol. abs/1807.03480, 2018. [Online].
  Available: \url{http://arxiv.org/abs/1807.03480}
\BIBentrySTDinterwordspacing

\bibitem{DBLP:conf/icra/NairBFLK20}
\BIBentryALTinterwordspacing
S.~Nair, M.~Babaeizadeh, C.~Finn, S.~Levine, and V.~Kumar, ``{TRASS:} time
  reversal as self-supervision,'' in \emph{2020 {IEEE} International Conference
  on Robotics and Automation, {ICRA} 2020, Paris, France, May 31 - August 31,
  2020}.\hskip 1em plus 0.5em minus 0.4em\relax {IEEE}, 2020, pp. 115--121.
  [Online]. Available: \url{https://doi.org/10.1109/ICRA40945.2020.9196862}
\BIBentrySTDinterwordspacing

\bibitem{10.5555/3295222.3295258}
M.~Andrychowicz, F.~Wolski, A.~Ray, J.~Schneider, R.~Fong, P.~Welinder,
  B.~McGrew, J.~Tobin, P.~Abbeel, and W.~Zaremba, ``Hindsight experience
  replay,'' in \emph{Proceedings of the 31st International Conference on Neural
  Information Processing Systems}, ser. NIPS'17.\hskip 1em plus 0.5em minus
  0.4em\relax Curran Associates Inc., 2017, pp. 5055--5065.

\bibitem{DBLP:journals/corr/abs-1909-05829}
\BIBentryALTinterwordspacing
S.~Nair and C.~Finn, ``Hierarchical foresight: Self-supervised learning of
  long-horizon tasks via visual subgoal generation,'' \emph{CoRR}, vol.
  abs/1909.05829, 2019. [Online]. Available:
  \url{http://arxiv.org/abs/1909.05829}
\BIBentrySTDinterwordspacing

\bibitem{10.5555/3298483.3298491}
P.-L. Bacon, J.~Harb, and D.~Precup, ``The option-critic architecture,'' ser.
  AAAI'17.\hskip 1em plus 0.5em minus 0.4em\relax AAAI Press, 2017, pp.
  1726--1734.

\bibitem{10.5555/3327144.3327250}
O.~Nachum, S.~Gu, H.~Lee, and S.~Levine, ``Data-efficient hierarchical
  reinforcement learning,'' in \emph{Proceedings of the 32nd International
  Conference on Neural Information Processing Systems}, ser. NIPS'18, 2018, pp.
  3307--3317.

\bibitem{eysenbach2018diversity}
\BIBentryALTinterwordspacing
B.~Eysenbach, A.~Gupta, J.~Ibarz, and S.~Levine, ``Diversity is all you need:
  Learning skills without a reward function,'' in \emph{International
  Conference on Learning Representations}, 2019. [Online]. Available:
  \url{https://openreview.net/forum?id=SJx63jRqFm}
\BIBentrySTDinterwordspacing

\bibitem{mitash2020scene}
C.~Mitash, B.~Wen, K.~Bekris, and A.~Boularias, ``Scene-level pose estimation
  for multiple instances of densely packed objects,'' in \emph{Conference on
  Robot Learning}.\hskip 1em plus 0.5em minus 0.4em\relax PMLR, 2020, pp.
  1133--1145.

\bibitem{morgan2021vision}
A.~S. Morgan, B.~Wen, J.~Liang, A.~Boularias, A.~M. Dollar, and K.~Bekris,
  ``Vision-driven compliant manipulation for reliable, high-precision assembly
  tasks,'' \emph{RSS}, 2021.

\bibitem{ho2016generative}
J.~Ho and S.~Ermon, ``Generative adversarial imitation learning,''
  \emph{Advances in neural information processing systems}, vol.~29, pp.
  4565--4573, 2016.

\bibitem{pomerleau1989alvinn}
D.~A. Pomerleau, ``Alvinn: An autonomous land vehicle in a neural network,''
  Carnegie-Mellon University Pittsburgh PA, Tech. Rep., 1989.

\bibitem{wen2021bundletrack}
B.~Wen and K.~Bekris, ``Bundletrack: 6d pose tracking for novel objects without
  instance or category-level 3d models,'' \emph{IROS}, 2021.

\end{thebibliography}
